\documentclass[twoside,11pt]{article}

\usepackage{subfigure}
\usepackage{paralist,amsmath, amssymb, bm,color}
\usepackage{algorithm, algorithmic}

\usepackage[preprint]{jmlr2e}

\usepackage{jmlr2e}
\usepackage{hyperref}
\def \y {\mathbf{y}}

\def \x {\mathbf{x}}
\def \g {\mathbf{g}}

\def \z {\mathbf{z}}
\def \u {\mathbf{u}}

\def \w {\mathbf{w}}
\def \R {\mathbb{R}}

\def \P {\mathcal{P}}

\def \A {\mathcal{A}}

\def \B {\mathbf{B}}

\def \wt {\widetilde{w}}

\def \B {\mathcal{B}}
\def \X {\mathcal{X}}

\def \Lt{\tilde{L}}

\def \lbah {\widehat{\lambda}}
\def \laha {\widehat{\alpha}}

\def \expc {exp}
\def \str {str}
\def \con {con}
\def \smo {smo}
\def \Ept {\mathcal{E}}

\def \ta        {\mathtt{term~(a)}}
\def \tb        {\mathtt{term~(b)}}
\def \tc        {\mathtt{term~(c)}}

\DeclareMathOperator*{\argmin}{argmin}

\newtheorem{assumption}{Assumption}

\makeatletter
\AtBeginDocument{\Hy@breaklinkstrue}
\makeatother

\usepackage{lastpage}
\jmlrheading{1}{2024}{1-\pageref{LastPage}}{7/24}{}{}{Lijun Zhang, Yibo Wang, Guanghui Wang, Jinfeng Yi, Tianbao Yang}

\ShortHeadings{Universal OCO Meets Second-order Bounds}{Zhang, Wang, Wang, Yi, Yang}
\firstpageno{1}

\begin{document}

\title{Universal Online Convex Optimization \\ Meets Second-order Bounds}

\author{\name Lijun Zhang \email zhanglj@lamda.nju.edu.cn 
        \AND
        \name Yibo Wang \email wangyb@lamda.nju.edu.cn \\
       \addr National Key Laboratory for Novel Software Technology,  Nanjing University,  China\\School of Artificial Intelligence, Nanjing University,  China
       \AND
       \name Guanghui Wang \email gwang369@gatech.edu \\
       \addr College of Computing, Georgia Institute of Technology, USA
       \AND
       \name Jinfeng Yi \email yijinfeng@jd.com \\
       \addr JD AI Research, Beijing, China
       \AND 
       \name Tianbao Yang \email tianbao-yang@tamu.edu\\
       \addr Department of Computer Science and Engineering, Texas A$\&$M University, USA
}

\editor{My editor}

\maketitle

\begin{abstract}%   <- trailing '%' for backward compatibility of .sty file
Recently, several universal methods have been proposed for online convex optimization, and attain minimax rates for multiple types of convex  functions simultaneously. However, they need to design and optimize one surrogate loss for each type of functions, making it difficult to exploit the structure of the problem and utilize existing algorithms. In this paper, we propose a simple strategy for universal online convex optimization, which avoids these limitations. The key idea is to construct a set of experts to process the \emph{original} online functions, and deploy a meta-algorithm over the \emph{linearized} losses to aggregate predictions from experts. Specifically, the meta-algorithm is required to yield a second-order bound with excess losses, so that it can leverage strong convexity and exponential concavity to control the meta-regret. In this way, our strategy inherits the theoretical guarantee of \emph{any} expert designed for strongly convex functions and exponentially concave functions, up to a double logarithmic factor. As a result, we can plug in off-the-shelf online solvers as black-box experts to deliver problem-dependent regret bounds. For general convex functions, it maintains the minimax optimality and also achieves a small-loss bound. 
Furthermore, we extend our universal strategy to online composite optimization, where the loss function comprises a time-varying function and a fixed regularizer. To deal with the composite loss functions, we employ a meta-algorithm based on the optimistic online learning framework, which not only possesses a second-order bound, but also can utilize estimations for upcoming loss functions.  With appropriate configurations, we demonstrate that the additional regularizer does not contribute to the meta-regret, thus maintaining the universality in the composite setting.

\end{abstract}


\section{Introduction}
Online convex optimization (OCO) has become a leading online learning framework, capable of modeling various real-world problems such as online routing and spam filtering \citep{Intro:Online:Convex}. OCO can be seen as a structured repeated game, with the following protocol. At each round $t$, the online learner chooses $\x_t$ from a convex set $\X$. After committing to this choice, a convex cost function $f_t:\X \mapsto \R$ is revealed, and the loss incurred by the learner is $f_t(\x_t)$. The learner aims to minimize the cumulative loss over $T$ rounds, i.e., $\sum_{t=1}^T f_t(\x_t)$, which is equivalent to minimizing the regret \citep{bianchi-2006-prediction}:
\begin{equation}\label{eqn:reg}
\sum_{t=1}^T f_t(\x_t) - \min_{\x \in \X} \sum_{t=1}^T f_t(\x)
\end{equation}
defined as the excess loss suffered by the learner compared to the minimum loss of any fixed choice.

In the literature, there are plenty of algorithms for OCO \citep{Online:suvery,Modern:Online:Learning}. For example, online gradient descent (OGD) with $O(1/\sqrt{t})$ step size achieves $O(\sqrt{T})$ regret for general convex functions \citep{zinkevich-2003-online}; OGD with $O(1/t)$ step size attains $O(\log T)$ regret for strongly convex functions \citep{ICML_Pegasos}; online Newton step (ONS) enjoys $O(d \log T)$ regret for exponentially concave (abbr.~exp-concave) functions, where $d$ is the dimensionality \citep{ML:Hazan:2007}. Besides, there exist more powerful online algorithms such as ADAGRAD \citep{JMLR:Adaptive} that are equipped with problem-dependent regret bounds \citep{NIPS2010_Smooth,Gradual:COLT:12,Adam,pmlr-v70-mukkamala17a,ICLR:2018:Adam}, which become tighter when the problem has special structures. Although we have rich theories for OCO, its application requires heavy domain knowledge: Users must know the type of functions in order to select an appropriate algorithm, and when dealing with strongly convex functions and exp-concave functions, they also need to estimate the moduli of strong convexity and exponential concavity.

The lack of universality of previous algorithms motivates the development of universal methods for OCO \citep{NIPS2007_3319,icml2009_033}. One milestone is  MetaGrad of \citet{NIPS2016_6268}, which can handle general convex functions as well as exp-concave functions. Later, \citet{Adaptive:Maler} propose Maler, which further supports strongly convex functions explicitly. In a subsequent work, \citet{AAAI:2020:Wang} develop UFO, which exploits smoothness to deliver small-loss regret bounds, i.e., regret bounds that depend on the minimal loss. However, the three aforementioned methods need to design one surrogate loss for each possible type of functions, which is both tedious and challenging. Furthermore, because they rely on surrogate losses, it is difficult to produce problem-dependent regret bounds, except the small-loss one.

To avoid the above limitations, we propose a simple yet universal strategy for online convex optimization, including both the standard setting where the learner only suffers a time-varying loss $f_t(\x_t)$ in each round, and the composite setting where the learner incurs a composite loss $f_t(\x_t) + r(\x_t)$ with an additional regularizer $r(\x)$.  The key idea   is to build a set of experts to process the \emph{original} online functions, and deploy a meta-algorithm, which ensures a second-order bound with excess losses,  to aggregate predictions from experts. Since the expert observes the original functions, it can utilize their structures to  deliver problem-dependent regret bounds. The second-order bound of the meta-algorithm facilitates the management for different types of online functions, securing the  universality of our strategy. In the following, we specify our strategy for the two settings.

\paragraph{Standard Online Convex Optimization.} In the standard setting,  we first create a set of experts to handle the uncertainty of the type of online functions and (possibly) the associated parameters. When facing unknown continuous variables, we discretize them by constructing a geometric series to cover the range of their values. Second, we run a meta-algorithm to track the best expert on the fly, but use the \emph{linearized} losses to measure the performance. To benefit from strong convexity and exponential concavity, we require the meta-algorithm to yield a second-order bound with excess losses, and choose Adapt-ML-Prod \citep{pmlr-v35-gaillard14} as an example. Specifically, let $\ell_t$ and $\ell_t^i$ be the loss of the meta-algorithm and the $i$-th expert in the $t$-th round, respectively. The regret of the meta-algorithm satisfies
\begin{equation} \label{eqn:second:order}
    \sum_{t=1}^T (\ell_t-\ell_t^i) = O \left( \sqrt{\sum_{t=1}^T (\ell_t-\ell_t^i)^2} \right), \ \forall i
\end{equation}
where for brevity we drop the dependence on the number of experts.

By incorporating existing methods for strongly convex functions, exp-concave functions and general convex functions as experts, we obtain a universal algorithm with the following properties.
\begin{compactitem}
    \item For strongly convex functions, our algorithm is agnostic to the modulus of strong convexity, at the price of maintaining $O(\log T)$ experts. More importantly, it inherits the regret bound of \emph{any} expert designed for strongly convex functions, with a negligible additive factor of $O(\log \log T)$. As a result, we can deliver any problem-dependent or independent regret bound, without prior knowledge of strong convexity.
    \item For exp-concave functions, the above statements are also true.
    \item For general convex functions, the theoretical guarantee is a mix of the regret bound of the expert and the second-order bound of Adapt-ML-Prod. When the functions are convex and smooth, it yields a small-loss bound.
\end{compactitem}

Compared to previous universal methods \citep{NIPS2016_6268,Adaptive:Maler,AAAI:2020:Wang}, our algorithm has the following advantages.
\begin{compactitem}
    \item It decouples the loss used by the expert-algorithm and that by the meta-algorithm. In this way, we can directly utilize existing online algorithms as black-box subroutines, and do not need to design surrogate losses.
    \item For strongly convex functions and exp-concave functions, the regret bound of our algorithm achieves \emph{the best of all worlds}, provided that both the domain and gradients are bounded.
\end{compactitem}

\paragraph{Online Composite Optimization.} In the composite setting,  the loss function is defined as the sum of two functions:
\begin{equation}    \label{eqn:com:f+r}
    f_t(\x)+r(\x)
\end{equation}
where $f_t(\x)$ is a time-varying function, and $r(\x)$ is a fixed convex regularizer, such as   the $\ell_1$-norm  and the trace norm  \citep{JRSS:1996:lasso,PJO:2010:Nuclear}. Consequently, our goal is to minimize the  regret in terms of \eqref{eqn:com:f+r}:
\begin{equation}    \label{eqn:com:regret}
\sum_{t=1}^T \big[f_t(\x_t) +r(\x_t) \big] - \min_{\x \in \X} \sum_{t=1}^T \big[ f_t(\x) + r(\x) \big].
\end{equation}
To minimize \eqref{eqn:com:regret}, a natural attempt is to treat the sum of $f_t(\x)+r(\x)$ as a new function, and pass it to our universal algorithm designed for the standard setting. However, this approach exhibits several limitations. For example, for an exp-concave function $f_t(\x)$ and a convex regularizer $r(\x)$, the summation function is not necessarily an exp-concave function \citep{AISTATS:2018:Yang}, so that the optimal regret for exp-concave loss functions is unattainable. Moreover, the summation function ignores the presence of the regularizer, consequently failing to leverage the benefit of  $r(\x)$, such as the sparsity introduced by the $\ell_1$-norm. Therefore, we need to modify the previous universal strategy to explicitly support composite functions.

In the composite setting, we first construct  a set of experts  in the same way  as the standard setting to process the \textit{original composite} loss functions. Then, we run our meta-algorithm on the \textit{composite linearized} loss, which includes the linearized time-varying function  and the regularier. Moreover, instead of utilizing Adapt-ML-Prod as the meta-algorithm, we choose  an optimistic online learning method, called  Optimistic-Adapt-ML-Prod \citep{NIPS2016_405e2890}, which not only guarantees a second-order bound, but also can exploit estimations for upcoming loss functions. With appropriate estimations, we demonstrate that Optimistic-Adapt-ML-Prod is able to yield the second-order bound that solely depends on the time-varying function $f_t(\x)$, so that the  strong convexity and exponential concavity of $f_t(\x)$ can be utilized to control the meta-regret. By employing previous methods in online composite optimization as experts, our universal algorithm maintains its favorable properties. Specifically,  it preserves the regret bound of \emph{any} expert designed for  strongly convex time-varying functions and exp-concave time-varying functions, up to a negligible  double logarithmic factor. As a result, we can still deliver any problem-dependent or independent regret bounds. For general convex time-varying functions, our method ensures the guarantee that comprises the regret bound of  the expert and the second-order bound of Optimistic-Adapt-ML-Prod.

\paragraph{New Regret Bounds.} 
During the analysis, we revisit three existing methods for standard online convex optimization and online composite optimization \citep{COLT:2010:Composite,Gradual:COLT:12,TAC:2023:Scroccaro}, and derive \textit{new} theoretical results as byproducts.

\begin{compactitem}    
    \item In the standard setting with smooth and strongly convex functions, we  extend the online extra-gradient descent (OEGD) of \cite{Gradual:COLT:12}, and  establish a novel gradient-variation bound (Theorem~\ref{thm:OEGD:Strong}), i.e.,~regret bounds that depend on the variation of $\nabla f_t(\x)$;

    \item In the composite setting with smooth and strongly convex time-varying functions, and smooth and general convex  time-varying functions, we re-analyze the  composite objective mirror descent (COMID)  of \cite{COLT:2010:Composite}, and establish new (pseudo) small-loss bounds (Theorems~\ref{thm:COMID:con}~and~\ref{thm:COMID:str}), i.e.,~regret bounds that depend on the cumulative loss on $f_t(\x)$ suffered by the algorithm itself;

    \item In the composite setting with  smooth and exp-concave time-varying functions, we generalize the optimistic composite mirror descent (OCMD) of \cite{TAC:2023:Scroccaro}, and  derive a new gradient-variation bound (Theorem~\ref{thm:OCMD:exp:VT}) and a new (pseudo) small-loss bound (Theorem~\ref{thm:OCMD:exp:LT}).

\end{compactitem}
Although the above results are not our primary focus, they further enrich the developments of online convex optimization.

A preliminary version of this paper was presented at the $39$th International Conference on Machine Learning \citep{ICML:2022:Zhang}. In this paper, we  expand upon the conference version by further investigating online composite optimization and proposing a new universal algorithm designed for the composite loss functions.  Moreover, two analysis byproducts in the composite setting also serve as the extensions.

\paragraph{Organization.} The remainder of the paper is structured as follows. Section~\ref{sec:related-work} briefly reviews the related work. Section~\ref{sec:standard-OCO} illustrates our main results for standard online convex optimization, including the specification of our universal strategy and its theoretical guarantees for  strongly convex,  exp-concave and general convex  functions, respectively. In Section~\ref{sec:composite-OCO}, we extend the investigations to online composite optimization. Section~\ref{sec:analysis} presents the analysis of all theorems.  Section~\ref{sec:conclusion} concludes this paper and discusses future work. The appendix details our revisitation as well as new theoretical results for previous methods.

\section{Related Work}
\label{sec:related-work}
In this section, we review the related work in OCO, including traditional algorithms, universal algorithms, online composite optimization, and parameter-free algorithms.

\subsection{Traditional Algorithms}
For general convex functions, the most popular algorithm is online gradient descent (OGD), which attains $O(\sqrt{T})$ regret by setting the step size as $\eta_t=O(1/\sqrt{t})$ \citep{zinkevich-2003-online}. For $\lambda$-strongly convex functions, the regret bound can be improved to $O(\frac{1}{\lambda} \log T)$ by running OGD with the step size $\eta_t=O(1/[\lambda t])$ \citep{ICML_Pegasos}. For $\alpha$-exp-concave functions, online Newton step (ONS), with prior knowledge of the parameter $\alpha$, achieves $O(\frac{d}{\alpha} \log T)$ regret, where $d$ is the dimensionality \citep{ML:Hazan:2007}. These regret bounds for general convex functions, strongly convex functions, and exp-concave functions are known to be minimax optimal \citep{Lower:bound:Portfolio,Minimax:Online}, which means that they cannot be improved in the worst case. However, these bounds only exhibit the relationship with problem-independent quantities, such as the time horizon $T$ and the dimensionality $d$, and thus do not reflect the property of the online problem at hand.

To exploit the structure of the problem, various problem-dependent (or data-dependent) regret bounds have been established in recent years \citep{NIPS2010_Smooth,COLT:Adaptive:Subgradient,JMLR:Adaptive,Gradual:COLT:12,Beyond:Logarithmic,tieleman2012lecture,ADADELTA,Gradual:ML:14,Adam,pmlr-v70-mukkamala17a,ICLR:2018:Adam,ICLR:2020:Wang}. The problem-dependent bounds reduce to the minimax rates in the worst case, but can be better under favorable conditions.

One well-known result is the small-loss bound which is very popular in the studies of online learning \citep{LITTLESTONE1994212,AUER200248,Shai:thesis,pmlr-v40-Luo15}. When the functions are smooth and nonnegative, the regret for general convex functions, $\lambda$-strongly convex functions, and $\alpha$-exp-concave functions can be upper bounded by $O(\sqrt{L_T^*})$, $O(\frac{1}{\lambda} \log L_T^*)$, and $O(\frac{d}{\alpha} \log L_T^*)$ respectively, where
\begin{equation}\label{eqn:L:Star}
    L_T^*=\min_{\x \in \X} \sum_{t=1}^T f_t(\x)
\end{equation}
is the cumulative loss of the best point in $\X$ \citep{NIPS2010_Smooth,Beyond:Logarithmic,Adaptive:Regret:Smooth:ICML,AAAI:2020:Wang}. These small-loss bounds could be much tighter when $L_T^*$ is small, and still ensure the minimax optimality otherwise. Another problem-dependent guarantee is the gradient-variation bound for smooth functions \citep{Gradual:COLT:12,Gradual:ML:14,AISTATS:2016:Mohri}, which replaces $L_T^*$ in the upper bounds with the gradient variation:
\begin{equation}\label{eqn:grad:var}
    V_T=\sum_{t=1}^T \max_{\x \in \X} \|\nabla f_t(\x) - \nabla f_{t-1}(\x)\|_2^2.
\end{equation}
Then, the regret bounds become smaller if the online functions evolve gradually.

Besides smoothness, it is also possible to exploit other structural properties of the functions, such as the sparsity of gradients. One representative work is ADAGRAD \citep{COLT:Adaptive:Subgradient,JMLR:Adaptive}, which incorporates knowledge of the geometry of the data observed in earlier iterations to perform more informative gradient-based learning.  Let $\g_{1:T,j}$ be the vector obtained by concatenating the $j$-th element of the gradient sequence $\nabla f_1(\x_1), \ldots, \nabla f_T(\x_T)$.  ADAGRAD achieves
\[
    O\left(\sum_{j=1}^d \|\g_{1:T,j}\|_2 \right) \textrm{ and } O\left(\frac{1}{\lambda} \sum_{j=1}^d \log \|\g_{1:T,j}\|_2 \right)
\]
regret for general convex functions and $\lambda$-strongly convex functions, respectively \citep{COLT:Adaptive:Subgradient}. These two bounds match the minimax rates in the worst case, and become tighter when gradients are sparse. Since the seminal work of ADAGRAD, a series of problem-dependent online algorithms have been developed, including RMSprop \citep{tieleman2012lecture,pmlr-v70-mukkamala17a}, Adam \citep{Adam,ICLR:2018:Adam} and AdamW \citep{AdamW:ICLR:2019}.

Although there exist abundant algorithms and theories for OCO, how to choose them in practice is a nontrivial task. To ensure good performance, we not only need to know the type of functions, but also need to estimate the moduli of strong convexity and exponential concavity. The requirement of human intervention restricts the application of OCO to real-world problems, and motivates the development of universal algorithms for OCO.

\subsection{Universal Algorithms}

The first universal method for OCO is adaptive online gradient descent (AOGD) \citep{NIPS2007_3319}, which interpolates between the $O(\sqrt{T})$ regret of general convex functions and the $O(\log T)$ regret of strongly convex functions automatically. However, AOGD needs to know the modulus of strong convexity in each round, and does not support exp-concave functions.  \citet{icml2009_033} develop a proximal extension of AOGD, but it suffers the same limitations.

The studies of universal methods are further advanced by the MetaGrad algorithm of \citet{NIPS2016_6268}, which adapts to a much broader class of functions, including general convex functions and exp-concave functions. Under the framework of learning with expert advice \citep{bianchi-2006-prediction}, MetaGrad is a two-layer algorithm consisting of a set of experts and a meta-algorithm. To handle exponential concavity, each expert minimizes one surrogate loss
\begin{equation} \label{eqn:metagrad:sur}
    \begin{split}
        \ell_{t,\eta}^{\expc}(\x)=-\eta (\x_t -\x)^\top \g_t + \eta^2  \left[(\x_t -\x)^\top \g_t\right]^2
    \end{split}
\end{equation}
parameterized by the step size $\eta$, where $\g_t=\nabla f_t(\x_t)$.  MetaGrad maintains $O(\log T)$ experts to minimize (\ref{eqn:metagrad:sur}) with different step sizes, and  combines their predictions with a meta-algorithm. In this way, it attains $O(\frac{d}{\alpha} \log T)$ regret for $\alpha$-exp-concave functions, without knowing the value of $\alpha$. At the same time, MetaGrad also achieves an $O(\sqrt{T \log \log T} )$ regret bound for general convex functions. Although we can treat strongly convex functions as exp-concave and obtain an $O(d \log T)$ regret bound, there exists an $O(d)$ gap from the minimax rate of strongly convex functions \citep{Minimax:Online}.

To deal with strongly convex functions, \citet{Adaptive:Maler} design another surrogate loss
\begin{equation} \label{eqn:strong:sur}
    \begin{split}
        \ell_{t,\eta}^{\str}(\x)=-\eta (\x_t -\x)^\top \g_t + \eta^2  G^2 \|\x_t -\x\|_2^2
    \end{split}
\end{equation}
where $G$ is an upper bound of the norm of gradients. Their proposed method, named as Maler, introduces additional $O(\log T)$ experts to optimize (\ref{eqn:strong:sur}), and obtains $O(\frac{1}{\lambda} \log T)$ regret for $\lambda$-strongly convex functions. Similar to MetaGrad, it gets rid of the priori knowledge of strong convexity.  \citet{Adaptive:Maler} further propose the following surrogate loss:
\begin{equation} \label{eqn:cov:sur}
    \begin{split}
        \ell_{t,\eta}^{\con}(\x)=-\eta (\x_t -\x)^\top \g_t + \eta^2  G^2 D^2
    \end{split}
\end{equation}
where $D$ is an upper bound of the diameter of $\X$, and obtain the optimal $O(\sqrt{T})$ regret for general convex functions. To exploit smoothness, \citet{AAAI:2020:Wang} propose the following surrogate loss
\begin{equation} \label{eqn:smooth:strong:sur}
    \begin{split}
        \ell_{t,\eta}^{\str,\smo}(\x)=-\eta (\x_t -\x)^\top \g_t + \eta^2  \|\g_t\|_2^2 \|\x_t -\x\|_2^2
    \end{split}
\end{equation}
for strongly convex and smooth functions, reuse the surrogate loss in (\ref{eqn:metagrad:sur}) for exp-concave and smooth functions, and introduce the following surrogate loss
\begin{equation} \label{eqn:smooth:cov:sur}
    \begin{split}
        \ell_{t,\eta}^{\con,\smo}(\x)=-\eta (\x_t -\x)^\top \g_t + \eta^2  \|\g_t\|_2^2 D^2
    \end{split}
\end{equation}
for convex and smooth functions.  Under the smoothness condition, their algorithm, namely UFO, achieves $O(\frac{1}{\lambda} \log L_T^*)$, $O(\frac{d}{\alpha} \log L_T^*)$, and $O(\sqrt{L_T^*})$ regret bounds for $\lambda$-strongly convex functions, $\alpha$-exp-concave functions, and general convex functions respectively, where $L_T^*$ is defined in (\ref{eqn:L:Star}).

Besides supporting more types of functions, MetaGrad was also extended to avoid the knowledge of the Lipschitz constant $G$. Specifically, \citet{pmlr-v99-mhammedi19a} first design a basic algorithm called MetaGrad+C, which requires an initial estimate of the Lipschitz constant, and then uses a restarting scheme to set this parameter online. In this way, the final algorithm, named as MetaGrad+L, adapts to the Lipschitz hyperparameter automatically. Furthermore, \citet{pmlr-v99-mhammedi19a} also remove the need to specify the number of rounds in advance, and \citet{JMLR:v22:20-1444} propose a refined version of MetaGrad+L, which only restarts the meta-algorithm but not the experts.

From the above discussion, we observe that state-of-the-art universal methods in the literature \citep{NIPS2016_6268,Adaptive:Maler,AAAI:2020:Wang} all rely on the construction of surrogate losses, which brings two issues.
\begin{compactenum}
    \item They have to design one surrogate loss for each possible type of functions, which is very challenging. That is because we need to ensure that the regret in terms of surrogate losses can be converted back to the regret of original functions.
    
    \item Since all the experts receive the surrogate losses instead of the original functions, it is difficult to exploit  the structure of the particular problem instance.  Of course, one could use problem-dependent algorithms to optimize the surrogate losses, but they may not share the same structure with the original functions. Except the small-loss bound \citep{AAAI:2020:Wang}, it is unclear how to generate other problem-dependent regret bounds.
\end{compactenum}

\subsection{Online Composite Optimization}
Online composite optimization aims to minimize the composite loss functions in \eqref{eqn:com:f+r} to generate decisions with certain favorable properties, such as the sparsity introduced by the $\ell_1$-norm  regularizer \citep{JRSS:1996:lasso} and the  low-rankness encouraged  by the trace norm regularizer \citep{PJO:2010:Nuclear}. 

In recent decades, there have been plenty of investigations in online composite optimization \citep{Others:2009:Duchi,NeurIPS:2009:Xiao,COLT:2010:Composite,Others:2023:Yang}. Specifically, the seminal work of \cite{Others:2009:Duchi}  proposes the forward backward splitting (FOBOS) method and establishes the $O(\sqrt{T})$ and $O( \frac{1}{\lambda} \log T)$ regret bounds for general convex and $\lambda$-strongly convex $f_t(\x)$, respectively. Later, based on the primal-dual subgradient framework \citep{Others:2009:Nesterov}, \cite{NeurIPS:2009:Xiao} develops the regularized dual averaging (RDA) method which ensures the same theoretical guarantees as FOBOS. \cite{COLT:2010:Composite} propose the composite objective mirror descent (COMID) method based on a different mirror descent framework \citep{Others:2002:Beck}, and also achieve the same $O(\sqrt{T})$ and $O( \frac{1}{\lambda} \log T)$ bounds for general convex and $\lambda$-strongly convex $f_t(\x)$, respectively. Recently, for $\alpha$-exp-concave $f_t(\x)$, \cite{Others:2023:Yang} propose the proximal online Newton step (ProxONS)  method and establish an $O(\frac{d}{\alpha} \log T)$ regret bound.

Besides the above methods, there are other efforts  investigating  problem-dependent regret bounds \citep{AISTATS:2016:Mohri,Others:2020:Joulani,TAC:2023:Scroccaro}, by exploiting the special properties of the problem, such as the smoothness of $f_t(\x)$. Specifically, based on the optimistic online learning framework \citep{COLT:2013:Rakhlin}, \cite{AISTATS:2016:Mohri}   introduce the adaptive optimistic FTRL (AO-FTRL) method, which is further generalized by \cite{Others:2020:Joulani}.  The key idea is to utilize an estimation of the upcoming loss function for decision updating. If the estimation is accurate, the regret can be very small; otherwise, the minimax rate is preserved. Based on similar ideas, \cite{TAC:2023:Scroccaro} develop the optimistic composite mirror descent (OptCMD) method, and  achieve   $O(\sqrt{V_T})$ and $O( \frac{1}{\lambda} \log V_T)$ regret bounds for smooth and general convex, and smooth and $\lambda$-strongly convex $f_t(\x)$, respectively.

Despite extensive methods in online composite optimization, applying them in practical problems still requires to know the type of loss functions in advance. Moreover, in the literature, there is no universal algorithm that explicitly handles composite losses.

\subsection{Parameter-Free Algorithms}
A parallel line of research is parameter-free online learning, which aims to design algorithms that avoid tuning parameters.  The motivation is that traditional online algorithms, such as OGD  \citep{zinkevich-2003-online}, require some prior knowledge about the comparator and/or gradients to set their parameters, e.g., the step size. To avoid this limitation,  parameter-free algorithms are designed to automatically adapt to both the norms of the comparator and  gradients \citep{NIPS2014_0ff8033c,NIPS2016_32072254,NIPS2016_550a141f,pmlr-v65-cutkosky17a,NIPS2017_a2186aa7,Scale:Free:Online,pmlr-v75-cutkosky18a,pmlr-v99-cutkosky19a,pmlr-v125-mhammedi20a}. The study of parameter-free algorithms mainly focuses on general convex functions, and is complementary to the development of universal algorithms.

\section{Standard Online Convex Optimization}
\label{sec:standard-OCO}
In this section, we first introduce necessary assumptions and definitions, and then explain the motivation of our paper. Next, we elaborate the proposed strategy for the standard OCO, and then present theoretical guarantees for strongly convex functions, exp-concave functions and general convex functions. Finally, we discuss parameter-free extensions. 

\subsection{Preliminaries}
We start with two common assumptions of OCO \citep{zinkevich-2003-online,NIPS2016_6268,Problem:Dynamic:Regret,Dual:Adaptivity}.

\begin{assumption}\label{ass:1} 
    The gradients of all functions are bounded by $G$, i.e.,
    \begin{equation}\label{eqn:grad}
        \max_{\x \in \X}\|\nabla f_t(\x)\|_2 \leq G, \ \forall t \in [T].
    \end{equation}
\end{assumption}

\begin{assumption}\label{ass:2} 
    The diameter of the domain $\X$ is bounded by $D$, i.e.,
    \begin{equation}\label{eqn:domain}
        \max_{\x, \y \in \X} \|\x -\y\|_2 \leq D.
    \end{equation}
\end{assumption}
To facilitate presentations, we assume the values of $G$ and $D$ are known beforehand, and will return to this issue in Section~\ref{sec:free}.

Then, we introduce the definitions of strongly convex functions and exp-concave functions \citep{Convex-Optimization,bianchi-2006-prediction}.

\begin{definition} \label{def:strong} 
    A function $f: \X \mapsto \R$ is $\lambda$-strongly convex if
    \begin{equation} \label{eqn:strong:convex:def}
        f(\y) \geq f(\x) +  \langle \nabla f(\x), \y -\x  \rangle + \frac{\lambda}{2} \|\y -\x \|_2^2,
    \end{equation}
    for all $\x, \y \in \X$.
\end{definition}

\begin{definition} \label{def:exp} 
    A function $f: \X \mapsto \R$ is $\alpha$-exp-concave if $\exp(-\alpha f(\cdot))$ is concave over $\X$.
\end{definition}
In the analysis, we will make use of the following property of exp-concave functions \citep[Lemma 3]{ML:Hazan:2007}.
\begin{lemma} \label{lem:exp} 
    Suppose $f:\X \mapsto \R$ is $\alpha$-exp-concave. Under Assumptions~\ref{ass:1} and \ref{ass:2},  we have
    \begin{equation} \label{eqn:exp:concave:prop}
        f(\y) \geq f(\x)+  \langle \nabla f(\x), \y -\x  \rangle + \frac{\beta}{2}   \langle \nabla f(\x), \y -\x  \rangle^2,
    \end{equation}
    for all $\x, \y \in \X$, where $\beta = \frac{1}{2} \min\{\frac{1}{4GD}, \alpha\}$.
\end{lemma}

\subsection{Motivations}
Similar to existing universal methods \citep{NIPS2016_6268,Adaptive:Maler,AAAI:2020:Wang}, we follow the framework of learning with expert advice. Specifically, we construct a set of experts for each type of functions and use a meta-algorithm to aggregate their predictions. The difference is that our universal strategy adopts two novel ingredients:
\begin{compactenum}[(i)]

    \item the experts process the \emph{original} functions so that they can exploit the structure of the problem instance to deliver problem-dependent regret bounds;
    
    \item  the meta-algorithm chooses the \emph{linearized} losses and makes use of second-order bounds to control the meta-regret.
\end{compactenum}
In the following, we take strongly convex functions as an example to explain the motivation.

To simplify notations, we assume that experts in the set $\Ept$ are ordered, and use $E^i$ to denote the $i$-th expert. Let $\x_t$ and $\x_t^i$ be the output of the meta-algorithm and the expert $E^i$ in the $t$-th round, respectively. 
The regret of the meta-algorithm can be decomposed as the sum of the meta-regret and the expert-regret:
\[
\begin{split}
     \sum_{t=1}^T    f_t(\x_t)- \min_{\x \in \X} \sum_{t=1}^T f_t(\x)=& \underbrace{\sum_{t=1}^T    f_t(\x_t)-\sum_{t=1}^T    f_t(\x_t^i)}_{:=\textrm{meta-regret}}+ \underbrace{\sum_{t=1}^T   f_t(\x_t^i) -  \min_{\x \in \X} \sum_{t=1}^T f_t(\x)}_{:=\textrm{expert-regret}}.
\end{split}
\]
To bound the expert-regret, we can directly utilize theoretical guarantees of the expert-algorithm. So, the key is  how to ensure a small meta-regret.

Instead of using the original function, our meta-algorithm chooses the linearized loss
\begin{equation}\label{eqn:linear:loss:1}
    l_t(\x)= \langle \nabla f_t(\x_t), \x - \x_t\rangle
\end{equation}
to measure the performance of the expert. From Definition~\ref{def:strong}, we have the following relationship between the meta-regret in terms of  $f_t(\cdot)$ and the meta-regret in terms of $l_t(\cdot)$:
\begin{equation} \label{eqn:meta:relation}
    \begin{split}
        \sum_{t=1}^T    f_t(\x_t)- \sum_{t=1}^T f_t(\x_t^i) \overset{ \eqref{eqn:strong:convex:def} }{\leq} &  \sum_{t=1}^T  \langle \nabla f_t(\x_t), \x_t -\x_t^i   \rangle  - \frac{\lambda}{2} \sum_{t=1}^T \| \x_t -\x_t^i\|_2^2 \\
        \overset{ \eqref{eqn:linear:loss:1} }{=} &  \sum_{t=1}^T  \big(l_t(\x_t) -l_t(\x_t^i)\big) - \frac{\lambda}{2} \sum_{t=1}^T \| \x_t -\x_t^i\|_2^2.
    \end{split}
\end{equation}

Since we require the meta-algorithm to yield a second-order bound in the form of (\ref{eqn:second:order}), the meta-regret in terms of $l_t(\cdot)$ becomes
\begin{equation} \label{eqn:second:linear:1}
    \begin{split}
        \sum_{t=1}^T  \big(l_t(\x_t) -l_t(\x_t^i)\big)\overset{\eqref{eqn:second:order}}{=} &O\left(  \sqrt{\sum_{t=1}^T \big(l_t(\x_t) -l_t(\x_t^i)\big)^2}\right)\\
        \overset{\eqref{eqn:linear:loss:1}}{=}& O\left(  \sqrt{\sum_{t=1}^T \langle \nabla f_t(\x_t), \x_t- \x_t^i \rangle^2}\right).
    \end{split}
\end{equation}
From Assumption~\ref{ass:1}, we further have
\begin{equation} \label{eqn:second:linear:2}
    \begin{split}
      \sum_{t=1}^T  \big(l_t(\x_t) -l_t(\x_t^i)\big) \overset{ \eqref{eqn:grad}, \eqref{eqn:second:linear:1}}{=} & O\left(  \sqrt{G^2\sum_{t=1}^T \|\x_t- \x_t^i \|_2^2}\right) \\
      =&O\left( \frac{G^2}{\lambda} \right) + \frac{\lambda}{2} \sum_{t=1}^T \| \x_t -\x_t^i\|_2^2.
    \end{split}
\end{equation}
Combining \eqref{eqn:meta:relation} and \eqref{eqn:second:linear:2}, we have the following meta-regret
\[
    \sum_{t=1}^T    f_t(\x_t)- \sum_{t=1}^T f_t(\x_t^i) = O\left( \frac{G^2}{\lambda} \right) 
\]
which is small and thus does not affect the optimality of the algorithm.

From the above discussions, we have the following conclusion: It is the second-order bound in (\ref{eqn:second:linear:1}) that makes it possible to exploit the negative term in (\ref{eqn:meta:relation}), which arises from the strong convexity. Based on Lemma~\ref{lem:exp}, we observe a similar phenomenon for exp-concave functions.

\subsection{Our Universal Strategy}
Our universal strategy for standard online convex optimization (USC) is summarized in Algorithm~\ref{alg:1}.  We will consider three types of convex functions: strongly convex functions, exp-concave functions, and general convex functions.

Let $\A_{\str}$ be the set of candidate algorithms designed for strongly convex functions, and $\P_{\str}$ be the set of possible values of the modulus of strong convexity. Note that although the modulus of strong convexity is continuous, we can construct a finite set $\P_{\str}$ to approximate its value, which will be elaborated in Section~\ref{sec:strong:convex}. For each algorithm $A \in \A_{\str}$ and each $\lambda \in \P_{\str}$, we create an expert $E(A,\lambda)$ by invoking algorithm $A$ with parameter $\lambda$, and add it to the set $\Ept$ consisting of experts (Steps 3-8). Similarly, let $\A_{\expc}$ be the set of algorithms designed for exp-concave functions, and $\P_{\expc}$ be the set of values of the modulus of exponential concavity. For each algorithm $A \in \A_{\expc}$ and each $\alpha \in \P_{\expc}$, we instantiate an expert $E(A,\alpha)$ by running algorithm $A$ with parameter $\alpha$, and add it to $\Ept$ (Steps 9-14). General convex functions can be handled more easily, and we denote the set of algorithms designed for them by $\A_{\con}$. Then, we simply create an expert $E(A)$ for each $A \in \A_{\con}$ and add it to $\Ept$ (Steps 15-18).

\begin{algorithm}[tb]
   \caption{A Universal Strategy for Online Convex Optimization (USC)}
   \label{alg:1}
\begin{algorithmic}[1]
   \STATE {\bfseries Input:}  $\A_{\str}$, $\A_{\expc}$, $\A_{\con}$, $\P_{\str}$ and $\P_{\expc}$ \vspace{.5ex}
    \STATE Initialize $\Ept=\emptyset$  \vspace{.5ex}

    \FOR{each algorithm $A \in \A_{\str}$}
     \FOR{each $\lambda \in \P_{\str}$}
     \STATE Create an expert $E(A,\lambda)$ % by invoking algorithm $A$ with parameter $\lambda$
     \STATE $\Ept=\Ept \cup E(A,\lambda)$ %Add the expert $E(A,\lambda)$ to the set \Ept
   \ENDFOR
   \ENDFOR\vspace{.5ex}

   \FOR{each algorithm $A \in \A_{\expc}$}
     \FOR{each $\alpha \in \P_{\expc}$}
     \STATE Create an expert $E(A,\alpha)$ %by invoking algorithm $A$ with parameter $\alpha$
     \STATE $\Ept=\Ept \cup E(A,\alpha)$ %Add the expert $E(A,\alpha)$ to the set $\Ept$
      \ENDFOR
   \ENDFOR\vspace{.5ex}

   \FOR{each algorithm $A \in \A_{\con}$}
     \STATE Create an expert $E(A)$ % by invoking algorithm $A$
     \STATE $\Ept=\Ept \cup E(A)$ %Add the expert $E(A)$ to the set $\Ept$
   \ENDFOR \vspace{.5ex}

   \FOR{$t=1$ {\bfseries to} $T$}
   \STATE Calculate the weight $p_t^i$ of each expert $E^i$ by (\ref{eqn:pt:definition})
   \STATE Receive $\x_t^i$ from each expert $E^i$ in $\Ept$
   \STATE  Output the weighted average $\x_t = \sum_{i=1}^{|\Ept|} p_t^i \x_t^i$
    \STATE Observe the loss function $f_t(\cdot)$
    \STATE Send the required information of $f_t(\cdot)$ to each expert in $\Ept$
    \ENDFOR
\end{algorithmic}
\end{algorithm}

Next, we deploy a meta-algorithm to track the best expert on the fly. Here, one can use any method that enjoys second-order bounds with excess losses  \citep{pmlr-v40-Koolen15a,pmlr-v99-mhammedi19a}. We choose Adapt-ML-Prod \citep{pmlr-v35-gaillard14} because it is more simple and already satisfies our requirements. In the $t$-th round, we denote by $p_t^i$ the weight assigned to $E^i$, and $\x_t^i$ the prediction of $E^i$. The weights are determined according to Adapt-ML-Prod (Step 20).  After receiving predictions from all experts in $\Ept$ (Step 21), USC submits the weighted average (Step 22):
\begin{equation} \label{eqn:meta:prediction}
    \x_t = \sum_{i=1}^{|\Ept|} p_t^i \x_t^i.
\end{equation}
Then, it observes the loss function $f_t(\cdot)$ and sends the required information to all experts so that they can update their predictions (Steps 23-24). If the expert is a first-order algorithm, we only need to send the gradient of $f_t(\cdot)$.  Thus, USC may query the gradient \emph{multiple} times in each round. It is worth to mention that allowing the online learner to observe multiple gradients does not affect the minimax rates in the full-information setting \citep{Minimax:Online}, where the leaner can observe the entire function. So, we can still use existing minimax rates to verify the optimality of USC.

Finally, we briefly explain how to calculate the weights of experts, i.e., $p_t^i$'s.  As we explained before, the meta-algorithm uses the linearized loss in  \eqref{eqn:linear:loss:1} to measure the performance of each expert. In particular, the loss of $E^i$ is given by
\[
    l_t^i:= l_t(\x_t^i)= \langle \nabla f_t(\x_t), \x_t^i - \x_t  \rangle.
\]
Under Assumptions~\ref{ass:1} and \ref{ass:2}, we have
\[
    |l_t^i| \leq \|\nabla f_t(\x_t)\|_2 \|\x_t^i - \x_t\|_2 \overset{ \eqref{eqn:grad}, \eqref{eqn:domain}}{\leq}   GD.
\]
Because Adapt-ML-Prod requires the loss to lie in $[0,1]$, we normalize $l_t^i$ in the following way
\begin{equation} \label{eqn:linear:loss}
    \ell_t^i= \frac{ \langle \nabla f_t(\x_t), \x_t^i- \x_t \rangle + GD}{2GD} \in [0,1] .
\end{equation}
Then, the loss of the meta-algorithm suffered in the $t$-th round becomes
\begin{equation} \label{eqn:meta:loss}
    \ell_t= \sum_{i=1}^{|\Ept|} p_t^i \ell_t^i \overset{\eqref{eqn:meta:prediction},  \eqref{eqn:linear:loss}}{=} \frac{1}{2}.
\end{equation}
According to Adapt-ML-Prod \citep{pmlr-v35-gaillard14}, the weight of expert $E^i$ is determined by
\begin{equation} \label{eqn:pt:definition}
    p_t^i = \frac{\eta_{t-1}^i w_{t-1}^i }{\sum_{j=1}^{|\Ept|} \eta_{t-1}^j w_{t-1}^j}
\end{equation}
where
\begin{align}
    &\eta_{t-1}^i =  \min\left\{ \frac{1}{2}, \sqrt{\frac{\ln |\Ept|}{1+\sum_{s=1}^{t-1} (\ell_s - \ell_s^i )^2 }} \right\}, \ t \geq 1, \label{eqn:adap:rate} \\
    &w_{t-1}^i= \Big(w_{t-2}^i \big(1 + \eta_{t-2}^i (\ell_{t-1} - \ell_{t-1}^i ) \big) \Big)^{\frac{\eta_{t-1}^i}{\eta_{t-2}^i}}, \ t \geq 2. \nonumber
\end{align}
In the beginning, we set  $w_0^i= 1/|\Ept|$. As indicated by (\ref{eqn:adap:rate}), \citet{pmlr-v35-gaillard14} use an adaptive way to set multiple time-varying learning rates.

\subsection{Strongly Convex Functions} \label{sec:strong:convex}
We present the regret bound of our strategy when encountering strongly convex functions. To apply USC in Algorithm~\ref{alg:1}, we need to specify $\A_{\str}$, the set of candidate algorithms, and $\P_{\str}$, the set of possible values of the modulus of strong convexity. To build $\A_{\str}$, we can utilize any existing algorithm for online strongly convex optimization, such as
\begin{compactitem}
    \item OGD for strongly convex functions (SC-OGD) \citep{ICML_Pegasos};
    
    \item ADAGRAD for strongly convex functions \citep{COLT:Adaptive:Subgradient};
    
    \item Online extra-gradient descent (OEGD) for strongly convex and smooth functions \citep{Gradual:COLT:12};
    
    \item SC-RMSProp \citep{pmlr-v70-mukkamala17a};
    
    \item SAdam \citep{ICLR:2020:Wang};
    
    \item S$^2$OGD for strongly convex and smooth functions \citep{AAAI:2020:Wang}.
\end{compactitem}
\citet{Gradual:COLT:12} only investigate OEGD for exp-concave functions and general convex functions, under the smoothness condition. In Appendix~\ref{sec:app:OEGD}, we extend OEGD to strongly convex functions and obtain a gradient-variation bound of order $O(\log V_T)$, which may be of independent interest.\footnote{Compared to our conference version \citep{ICML:2022:Zhang}, we adopt a simpler step size from \citet[\S 4.3]{ICML:2023:Chen}.}

We proceed to construct $\P_{\str}$. Without loss of generality, we assume the unknown modulus $\lambda$ is both lower bounded and upper bounded. In particular, we assume $\lambda \in [1/T,1]$, because there is no need to explicitly consider the cases that $\lambda <1/T$ and $\lambda > 1$, as explained below.
\begin{compactenum}
    \item The regret bound of strongly convex functions exhibits an inverse dependence on $\lambda$. Thus, if $\lambda < 1/T$, the bound becomes at least $\Omega(T)$, which is meaningless. In this case, we cannot benefit from strong convexity and should treat these functions as general convex.
    
    \item From Definition~\ref{def:strong}, we know that $\lambda$-strongly convex functions with $\lambda >1$ are also $1$-strongly convex. So, they can be handled as $1$-strongly convex, and the resulting bound is optimal up to a constant (i.e., $\lambda$) factor.
\end{compactenum}

Based on the interval $[1/T,1]$, we set $\P_{\str}$ to be an exponentially spaced grid with a ratio of $2$:
\begin{equation} \label{eqn:discrete:lambda}
    \P_{\str}=\left\{\frac{1}{T}, \frac{2}{T}, \frac{2^2}{T}, \cdots, \frac{2^N}{T}  \right\}, \  N= \lceil\log_2 T \rceil.
\end{equation}
$\P_{\str}$ can approximate $\lambda$ well in the sense that for any $\lambda \in [1/T,1]$, there must exist a $\lbah \in \P_{\str}$ such that $\lbah \leq \lambda \leq 2 \lbah$.

In the following, we denote by $R(A,\lbah)$ the regret bound, predicted by theory, of expert $E(A,\lbah)$ in Algorithm~\ref{alg:1}. Note that the expert $E(A,\lbah)$ assumes the online functions are $\lbah$-strongly convex, which is true since $\lbah \leq \lambda$. Thus, the regret bound $R(A,\lbah)$ is \emph{valid}, and it is also \emph{tight} because $\lambda \leq 2 \lbah$. We have the following theoretical guarantee. % of USC.
\begin{theorem} \label{thm:USC:strong} 
    Under Assumptions \ref{ass:1} and \ref{ass:2}, if the online functions are $\lambda$-strongly convex with $\lambda \in [1/T, 1]$, Algorithm~\ref{alg:1} satisfies
    \[
        \begin{split}
            \sum_{t=1}^T f_t(\x_t) - \min_{\x \in \X} \sum_{t=1}^T f_t(\x) \leq &\min_{A \in \A_{\str}}  R(A,\lbah)+  2\Gamma GD \left(2 + \frac{1}{\sqrt{\ln |\Ept|}} \right) + \frac{\Gamma^2 G^2}{2 \lambda \ln |\Ept|}\\
             =& \min_{A \in \A_{\str}}  R(A,\lbah)+  O \left( \frac{\log \log T}{\lambda} \right)
        \end{split}
    \]
    where $\lbah \in \P_{\str}$, $\lbah \leq \lambda \leq 2 \lbah$, and
    \begin{equation} \label{eqn:beta:definiton}
        \begin{split}
            \Gamma  =  3\ln |\Ept| + \ln \left( 1 + \frac{|\Ept|}{2 e} \big(1 + \ln (T+1)\big)\right)  \overset{\eqref{eqn:number:experts}}{=}  O(\log \log T).
        \end{split}
    \end{equation}
\end{theorem}
\textbf{Remark:} To reveal the order of the upper bound, we assume the number of candidate algorithms is small, so $|\A_{\str}|$, $|\A_{\expc}|$ and $|\A_{\con}|$ are all small constants. Thus,
\begin{equation} \label{eqn:number:experts}
    \begin{split}
        |\Ept| =|\A_{\str}| \cdot |\P_{\str}|  + |\A_{\expc}| \cdot |\P_{\expc}| + |\A_{\con}|   \overset{\eqref{eqn:discrete:lambda},\eqref{eqn:discrete:alpha}}{=} O(\log T)
    \end{split}
\end{equation}
which is used in (\ref{eqn:beta:definiton}). When both the domain and gradients are bounded, Theorem~\ref{thm:USC:strong} shows that USC achieves \emph{the best of all worlds} for strongly convex functions, up to an additive factor of $O(\log \log T)$.

\noindent
\textbf{Remark:}  The computational complexity of an expert per iteration is generally independent from $T$,  but may depend on the dimensionality $d$ \citep{COLT:Adaptive:Subgradient}. To simplify discussions, we hide the dependence on $d$, and assume the complexity is $O(1)$ per iteration. Since USC maintains $|\Ept|=O(\log T)$ experts, its computational complexity is $O(\log T)$ per iteration, which is the same as that of previous  methods \citep{NIPS2016_6268,Adaptive:Maler,AAAI:2020:Wang}.

To be more concrete, we use the small-loss bound and the gradient-variation bound for smooth functions to give an example. To this end, we need an additional assumption \citep{NIPS2010_Smooth}.

\begin{assumption}\label{ass:3} 
    All the online functions are nonnegative, and $H$-smooth over $\X$.
\end{assumption}
By using OEGD \citep{Gradual:COLT:12} and S$^2$OGD \citep{AAAI:2020:Wang} as experts, we have the following corollary.
\begin{corollary} \label{cor:1}  Under Assumptions \ref{ass:1}, \ref{ass:2}, and \ref{ass:3}, if the online functions are $\lambda$-strongly convex with $\lambda \in [1/T, 1]$, we have
\[
    \begin{split}
        \sum_{t=1}^T f_t(\x_t) - \min_{\x \in \X} \sum_{t=1}^T f_t(\x)=O \left( \frac{1}{\lambda} \Big(\min(\log L_T^*, \log V_T) + \log\log T \Big) \right)
    \end{split}
\]
where $L_T^*$ and $V_T$ are defined in (\ref{eqn:L:Star}) and (\ref{eqn:grad:var}) respectively, provided $\textrm{OEGD}, \textrm{S}^2\textrm{OGD} \in \A_{\str}$.
\end{corollary}

\subsection{Exp-concave Functions}    \label{sec:exp:concave}
We move to exp-concave functions, and use the following algorithms to build $\A_{\expc}$:
\begin{compactitem}
    \item Online Newton step (ONS) \citep{ML:Hazan:2007};
    
    \item ONS for exp-concave and smooth functions \citep{Beyond:Logarithmic};
    
    \item OEGD for exp-concave and smooth functions \citep{Gradual:COLT:12}.
\end{compactitem}
Following the same arguments as in Section~\ref{sec:strong:convex}, we also assume the modulus of exponential concavity $\alpha$ lies in $[1/T,1]$, and use the same geometric series to construct $\P_{\expc}$ as
\begin{equation} \label{eqn:discrete:alpha}
    \P_{\expc}=\left\{\frac{1}{T}, \frac{2}{T}, \frac{2^2}{T}, \cdots, \frac{2^N}{T}  \right\}, \  N= \lceil\log_2 T \rceil.
\end{equation}
Then, for any $\alpha \in [1/T,1]$, there must exist an $\laha \in \P_{\expc}$ such that $\laha \leq \alpha \leq 2 \laha$.

We denote by $R(A,\laha)$ the regret bound of expert $E(A,\laha)$ in Algorithm~\ref{alg:1}. Similarly, $R(A,\laha)$ is both valid and tight. We have the following guarantee for exp-concave functions, which is analogous to Theorem~\ref{thm:USC:strong}.
\begin{theorem} \label{thm:USC:exp} 
    Under Assumptions \ref{ass:1} and \ref{ass:2}, if the online functions are $\alpha$-exp-concave with $\alpha \in [1/T, 1]$, Algorithm~\ref{alg:1} satisfies
    \[
        \begin{split}
            \sum_{t=1}^T f_t(\x_t) - \min_{\x \in \X} \sum_{t=1}^T f_t(\x) \leq & \min_{A \in \A_{\expc}}  R(A,\laha)+  2\Gamma GD \left(2 + \frac{1}{\sqrt{\ln |\Ept|}} \right) + \frac{\Gamma^2 }{2 \beta \ln |\Ept|}\\
            =& \min_{A \in \A_{\expc}}  R(A,\laha)+  O \left( \frac{\log \log T}{\alpha} \right)
        \end{split}
    \]
    where $\laha \in \P_{\expc}$, $\laha \leq \alpha \leq 2 \laha$, $\beta = \frac{1}{2} \min\{\frac{1}{4GD}, \alpha\}$, and $\Gamma$ is defined in (\ref{eqn:beta:definiton}).
\end{theorem}
\textbf{Remark:} Similar to the case of strongly convex functions, USC inherits the regret bound of \emph{any} expert designed for exp-concave functions, with a negligible double logarithmic factor. By using  ONS \citep{Beyond:Logarithmic} and OEGD \citep{Gradual:COLT:12} as experts, we obtain the best of the small-loss bound and the gradient-variation bound, up to a double logarithmic factor.

\begin{corollary} \label{cor:2}  
    Under Assumptions \ref{ass:1}, \ref{ass:2}, and \ref{ass:3}, if the online functions are $\alpha$-exp-concave with $\alpha \in [1/T, 1]$, we have
    \[
        \begin{split}
            \sum_{t=1}^T f_t(\x_t) - \min_{\x \in \X} \sum_{t=1}^T f_t(\x)=O \left( \frac{1}{\alpha} \Big(d \min(\log L_T^*, \log V_T) + \log\log T \Big) \right)
        \end{split}
    \]
    where $L_T^*$ and $V_T$ are defined in (\ref{eqn:L:Star}) and (\ref{eqn:grad:var}) respectively, provided $\textrm{ONS}, \textrm{OEGD} \in \A_{\expc}$.
\end{corollary}

\subsection{General Convex Functions}
Finally, we study general convex functions, and in this case, we have various algorithms to construct $\A_{\con}$, such as OGD \citep{zinkevich-2003-online}, ADAGRAD \citep{JMLR:Adaptive}, OEGD for convex and smooth functions \citep{Gradual:COLT:12}, RMSprop \citep{tieleman2012lecture}, ADADELTA \citep{ADADELTA}, Adam \citep{Adam}, AO-FTRL \citep{AISTATS:2016:Mohri}, and SOGD \citep{Adaptive:Regret:Smooth:ICML}.

Let $R(A)$ be the regret bound of expert $E(A)$ in Algorithm~\ref{alg:1}. The theoretical guarantee of USC for general convex functions is stated below.

\begin{theorem} \label{thm:USC:convex} 
    Under Assumptions \ref{ass:1} and \ref{ass:2}, for any sequence of convex functions, Algorithm~\ref{alg:1} satisfies
    \[
        \begin{split}
            \sum_{t=1}^T f_t(\x_t) - \min_{\x \in \X} \sum_{t=1}^T f_t(\x) \leq &\min_{A \in \A_{\con}}  R(A) +4 \Gamma GD   +   \frac{\Gamma D}{\sqrt{\ln |\Ept|}} \sqrt{4G^2 + \sum_{t=1}^T  \| \nabla f_t(\x_t) \|_2^2} \\
             =& \min_{A \in \A_{\con}}  R(A)+  O \left( \sqrt{T \log\log T} \right)
        \end{split}
    \]
where $\Gamma$ is defined in (\ref{eqn:beta:definiton}).
\end{theorem}
\textbf{Remark:} The above theorem is weaker than those for strongly convex functions and exp-concave functions. That is because we cannot eliminate the regret of the meta-algorithm, and the final regret is the sum of the expert-regret and the meta-regret. Nevertheless, Theorem~\ref{thm:USC:convex} still implies a small-loss bound for smooth functions, when SOGD \citep{Adaptive:Regret:Smooth:ICML} is used as the expert.

\begin{corollary} \label{cor:3}  
    Under Assumptions \ref{ass:1}, \ref{ass:2}, and \ref{ass:3}, for any sequence of convex functions, we have
    \[
        \sum_{t=1}^T f_t(\x_t) - \min_{\x \in \X} \sum_{t=1}^T f_t(\x) = O\left( \sqrt{L_T^* \log\log T} \right)
    \]
    where $L_T^*$ is defined in (\ref{eqn:L:Star}), provided $\textrm{SOGD} \in \A_{\con}$.
\end{corollary}

\subsection{Parameter-Free Extensions} \label{sec:free}
We note that USC requires the prior knowledge of the  gradient norm  bound  $G$ and the domain bound $D$, which are not always readily available in practical applications and thus motivate  us to investigate a parameter-free extension with unknown $G$ and $D$. Generally speaking, estimating $D$ is relatively straightforward since it is solely tied to the domain, whereas evaluating $G$ can be challenging, as it depends on all online functions. Therefore, in the following, we focus on how to avoid prior knowledge of $G$, provided that $G$ is bounded.

First, in Algorithm~\ref{alg:1}, $G$ is employed to normalize  the linearized loss in  \eqref{eqn:linear:loss}, enabling the application of the meta-algorithm Adapt-ML-Prod. When $G$  is unknown, we can opt for  more advanced methods which not only deliver second-order bounds but also adapt to the unknown loss range automatically, e.g., Squint+L \citep{pmlr-v99-mhammedi19a}. Consequently,  we can directly use the (unnormalized) linearized loss in \eqref{eqn:linear:loss:1}, maintaining the same guarantee for meta-regret.

Second, in some experts of Algorithm~\ref{alg:1}, $G$ is utilized to tune their parameters. When $G$ is unknown, we can select  experts that do not need to know the value of $G$. Specifically, for strongly convex functions, we can use OGD with the step size $\eta_t=1/(\lambda t)$ \citep{ICML_Pegasos} and AOGD \citep{NIPS2007_3319}. For exp-concave functions, we can choose the exponentially weighted online optimization (EWOO) algorithm \citep{ML:Hazan:2007} and MetaGrad+L \citep{pmlr-v99-mhammedi19a}. For general convex functions, we can use scale-free online learning \citep{Scale:Free:Online} and FreeGrad \citep{pmlr-v125-mhammedi20a}.

In summary,   due to the flexibility of USC, we can substitute both the meta-algorithm and experts with parameter-free alternatives that are capable of accommodating the unknown gradient norm bound  $G$, while preserving the universality of USC.

\section{Online Composite Optimization}
\label{sec:composite-OCO}
In this section, we extend our universal strategy to online composite optimization.

\subsection{Preliminaries}

First, we introduce the following standard assumptions in online composite optimization \citep{Others:2009:Duchi,COLT:2010:Composite}.

\begin{assumption}\label{ass:4} 
    The regularization function $r(\cdot)$ is convex over $\X$.
\end{assumption}

\begin{assumption}\label{ass:5} 
    The regularization function $r(\cdot)$ is non-negative and upper bounded by a known constant $C$, i.e., $\forall \x \in \X$, $0 \leq r(\x) \leq C$.
\end{assumption}

We assume that in the composite loss function \eqref{eqn:com:f+r}, $r(\cdot)$ is convex and known, but the function $f_t(\cdot)$ could be general convex, strongly convex, or exp-concave. In the literature, there have been plenty of studies for each specific setting  \citep{Others:2009:Duchi,NeurIPS:2009:Xiao,COLT:2010:Composite,Others:2023:Yang}, but all of them require the prior knowledge about $f_t(\cdot)$. In this paper, we aim to design a universal algorithm for online composite optimization, and attain the optimal regret for each possible case. To this end, one may attempt to treat the sum of $f_t(\x)+r(\x)$ as a new function, and pass it to USC. However, this approach introduces following issues. First, the sum of an exp-concave function and a convex function is not necessarily an exp-concave function \citep{AISTATS:2018:Yang}, implying that USC cannot deliver the optimal regret when $f_t(\cdot)$ is exp-concave.  Second, due to the additional component $r(\cdot)$, the summation function $f_t(\x)+r(\x)$ may lose certain special properties of $f_t(\cdot)$, e.g., the smoothness which is essential for delivering many problem-dependent bounds. Third, ignoring the composite structure  fails to harness the power of the regularizer. For instance, it is common to set $r(\cdot)$ as a sparsity-promoting regularizer, but treating the summation as a whole does not yield sparse solutions. Therefore, additional modifications are necessary to handle composite loss functions.

\subsection{Motivations}
Our universal algorithm for online composite optimization (USC-Comp) is still based on  the framework of learning with expert advice---building a set of experts for each type of functions and employing a meta-algorithm    to aggregate their predictions. Similar to USC, the experts are running over the original composite functions so that the  properties of the loss functions can be utilized to establish problem-dependent regret bounds. To handle the composite losses, we  modify  the meta-algorithm in the following way:

\begin{compactenum}[(i)]

    \item we run the meta-algorithm over a \textit{composite linearized} loss, which combines the  linearized loss and the regularization function, i.e.,
    \begin{equation}  
        l_t(\x):=  \langle \nabla f_t(\x_t), \x - \x_t  \rangle + r(\x)        
        \nonumber
    \end{equation}
    to measure the performance of experts;

    \item instead of employing Adapt-ML-Prod as the meta-algorithm, we utilize a different algorithm, called Optimistic-Adapt-ML-Prod \citep{NIPS2016_405e2890}, which, with appropriate configurations, can deliver a second-order bound that exclusively depends on the time-varying functions $f_t(\cdot)$ so that the strong convexity and exponential concavity of $f_t(\cdot)$ can be utilized to control the meta-regret.
\end{compactenum}

To make it clearer, we  take the exponential concavity case, i.e.,~$f_t(\cdot)$ is exp-concave,  as an example. In this case, the meta-algorithm should yield a second-order regret bound that solely depends on $f_t(\cdot)$. In other words, even though the regularization function $r(\cdot)$ is present in the meta-regret, it should not contribute to the upper bound. This is achievable due to the following fact:
\begin{quote}
  Prior to determining the weights for the experts, the meta-algorithm has already observed the part of the loss function that is related to the regularizer.
\end{quote}
The above fact  facilitates the possibility to set suitable configurations of  Optimistic-Adapt-ML-Prod to eliminate the influence of $r(\cdot)$ on the meta-regret.

Specifically, let $\x_t$ and $\x_t^i$ be the output of the meta-algorithm and an expert $E^i$ in the $t$-th round, respectively. Then, the regret in \eqref{eqn:com:regret} can be decomposed as the sum of the meta-regret and the expert-regret:
\begin{equation*}
    \begin{split}
         &\sum_{t=1}^T \big[f_t(\x_t) +r(\x_t) \big] - \min_{\x \in \X} \sum_{t=1}^T \big[ f_t(\x) + r(\x) \big] \\
         =& \underbrace{\sum_{t=1}^T  \left [ f_t(\x_t) + r(\x_t) \right]- \sum_{t=1}^T \left [ f_t(\x_t^i) + r(\x_t^i) \right ]}_{:=\textrm{meta-regret}}+ \underbrace{\sum_{t=1}^T  \left [ f_t(\x_t^i) + r(\x_t^i) \right]- \min_{\x \in \X} \sum_{t=1}^T \left [ f_t(\x) + r(\x) \right ]  }_{:=\textrm{expert-regret}}.
    \end{split}
\end{equation*}
The expert-regret can be upper bounded by leveraging theoretical results of existing methods for  online composite learning. Therefore, we only need to focus on the meta-regret.

To deal with the regularizer $r(\cdot)$, the meta-algorithm exploits the composite linearized loss 
\begin{equation}     \label{eqn:com:linear:loss}
    l_t^i:= l_t(\x_t^i) = \langle \nabla f_t(\x_t), \x_t^i - \x_t  \rangle + r(\x_t^i) 
\end{equation}
to measure the performance of $E^i$, where $\x_t = \sum_{i=1}^{|\Ept|} p_t^i \x_t^i$ denotes the decision made by the meta-algorithm and $p_t^i$ denotes the weight assigned to the expert $E^i$.  With the above composite linearized loss, the meta-regret is upper bounded by 
\begin{equation}    \label{eqn:com:meta-regret} 
    \begin{split}
        & \textrm{meta-regret}  \leq  \sum_{t=1}^T  \left[f_t(\x_t) - f_t(\x_t^i) \right]  + \sum_{t=1}^T \left[ \sum_{i = 1}^{|\Ept|} p_t^i r(\x_t^i) -     r(\x_t^i) \right]  \\
         & \overset{\eqref{eqn:exp:concave:prop}}{\leq}    \sum_{t=1}^T   \langle \nabla f_t(\x_t), \x_t- \x_t^i \rangle -   \frac{\beta}{2} \sum_{t=1}^T  \langle \nabla f_t(\x_t),  \x_t -\x_t^i \rangle^2   + \sum_{t=1}^T \left[ \sum_{i = 1}^{|\Ept|} p_t^i r(\x_t^i) -     r(\x_t^i) \right]   \\
         & \overset{\eqref{eqn:com:linear:loss}}{=}   \sum_{t=1}^T \left( \sum_{i=1}^{|\Ept|}  p_t^i l_t^i - l_t^i \right) -   \frac{\beta}{2} \sum_{t=1}^T  \langle \nabla f_t(\x_t),  \x_t -\x_t^i \rangle^2 \\
         & =   \sum_{t=1}^T \hat{R}_t^i -   \frac{\beta}{2} \sum_{t=1}^T  \langle \nabla f_t(\x_t),  \x_t -\x_t^i \rangle^2, 
    \end{split}
\end{equation}
where the first step is due to Jensen's inequality, i.e., $ r(\x_t) =   r( \sum_{i=1}^{|\Ept|} p_t^i \x_t^i)  \leq  \sum_{i=1}^{|\Ept|}  p_t^i r(\x_t^i) $, and for brevity, we define  
\begin{equation}      \label{eqn:com:R:hat}
    \hat{R}_t^i =   \sum_{i=1}^{|\Ept|}  p_t^i l_t^i - l_t^i = \langle \nabla f_t(\x_t), \x_t- \x_t^i \rangle    +   \sum_{i=1}^{|\Ept|}  p_t^i r(\x_t^i)  -  r(\x_t^i).    
\end{equation}
Note that  to manage the meta-regret, it is essential to constrain $\sum_{t=1}^T \hat{R}_t^i$ by a second-order bound which only depends on $f_t(\cdot)$, so that the negative term in \eqref{eqn:com:meta-regret} can be exploited.

To this end, we employ  Optimistic-Adapt-ML-Prod as the meta-algorithm, which utilizes an optimistic estimation (also called optimism)  to update $p_t^i$ for each expert $E^i$. In our algorithm,  we set the estimation as
\begin{equation}      \label{eqn:com:m:hat}
    \hat{m}_t^i =  \sum_{i=1}^{|\Ept|}  p_t^i r(\x_t^i) - r(\x_t^i).
\end{equation}
It should be emphasized that although $\hat{m}_t^i$ relies on $p_t^i$ which in turn depends on $\hat{m}_t^i$, we can still compute $\hat{m}_t^i$ in each round $t$ before updating $p_t^i$. The reasons lie in that (i) the regularizer term $r(\cdot)$ is fixed and has been already revealed to the meta-algorithm; (ii) the decision $\x_t^i$ can be sent to the meta-algorithm before computing $\hat{m}_t^i$; (iii) $\sum_{i=1}^{|\Ept|}  p_t^i r(\x_t^i)$ can be approximated efficiently  with little sacrifice in meta-regret bounds, which will be elaborated in Section~\ref{sec:com:universal}.

Then, according to Optimistic-Adapt-ML-Prod \citep[Theorem 3.4]{NIPS2016_405e2890}, we have 
\begin{equation} \label{eqn:com:pseudo:regret}
    \begin{split}
          \sum_{t=1}^T \hat{R}_t^i = & O\left( \sqrt{\sum_{t=1}^T \left(\hat{R}_t^i - \hat{m}_t^i \right)^2} \right) \overset{\eqref{eqn:com:R:hat}, \eqref{eqn:com:m:hat}}{=}   O\left(   \sqrt{\sum_{t=1}^T \langle \nabla f_t(\x_t), \x_t- \x_t^i \rangle^2}\right)  \\
          = & O\left( \frac{1}{\beta  }\right) +   \frac{\beta }{ 2  } \sum_{t=1}^T \langle \nabla f_t(\x_t), \x_t- \x_t^i \rangle^2.
    \end{split}
\end{equation}
Substituting  \eqref{eqn:com:pseudo:regret} into \eqref{eqn:com:meta-regret}, we obtain the following meta-regret bound:
\begin{equation}
    \begin{split}
        & \sum_{t=1}^T  \left [ f_t(\x_t) + r(\x_t) \right]- \sum_{t=1}^T \left [ f_t(\x_t^i) + r(\x_t^i) \right ] = O\left( \frac{1}{\beta }\right),
    \end{split}
\end{equation}
which is small and thus does not affect the overall regret bound of the algorithm. Similar results can also be obtained when   $f_t(\cdot)$ is strongly convex in \eqref{eqn:com:f+r}.

\subsection{Our Universal Strategy}
\label{sec:com:universal}

\begin{algorithm}[tb]
   \caption{A Universal Strategy for Online Composite Optimization (USC-Comp)}
   \label{alg:2}
    \begin{algorithmic}[1]
       \STATE {\bfseries Input:}  $\A_{\str}$, $\A_{\expc}$, $\A_{\con}$, $\P_{\str}$ and $\P_{\expc}$ \vspace{.5ex}
        \STATE Initialize $\Ept$ by following Steps $3$-$18$ in Algorithm~\ref{alg:1}  \vspace{.5ex}
        \STATE Observe the regularization term $r(\cdot)$
       \vspace{.5ex}
    
       \FOR{$t=1$ {\bfseries to} $T$}
       \STATE Receive $\x_t^i$ from each expert $E^i$ in $\Ept$
       \STATE Compute the optimism $m_t^i$ of each expert $E^i$ by \eqref{eqn:com:m}
       \STATE Calculate the weight $p_t^i$ of each expert $E^i$ by \eqref{eqn:com:pt:definition}
       \STATE  Output the weighted average $\x_t = \sum_{i=1}^{|\Ept|} p_t^i \x_t^i$
        \STATE Observe the loss function $f_t(\cdot)$
        \STATE Send the required information of $f_t(\cdot)$ and $r(\cdot)$ to each expert in $\Ept$
        \ENDFOR
    \end{algorithmic}
\end{algorithm}

We summarize our universal strategy for online composite optimization (USC-Comp) in Algorithm~\ref{alg:2}. Our new strategy is designed for three cases, in which the  regularization term $r(\cdot)$ is convex  and $f_t(\cdot)$ could be  strongly convex, exp-concave, or general convex. 

To handle the above cases simultaneously, we build the expert set $\Ept$ based on three candidate algorithm sets (i.e., $\A_{str}$, $\A_{exp}$ and $\A_{con}$),  each of which deals with  one case, and two modulus value sets (i.e., $\P_{str}$ and $\P_{exp}$), each of which contains possible values of the modulus of strong convexity or exponential concavity. The process of constructing $\Ept$ follows UCS in Algorithm~\ref{alg:1}. First, we initialize $\Ept$ with experts designed for the strongly convex function  $f_t(\cdot)$. Specifically, for each algorithm $A \in \A_{str}$ and each $\lambda \in \P_{str}$, we create an expert $E(A, \lambda)$ by running the algorithm $A$ with the parameter $\lambda$, and add it to the expert set $\Ept$. Then, for the exp-concave function  $f_t(\cdot)$ and general convex function  $f_t(\cdot)$, we repeat similar steps, adding an expert $E(A, \alpha)$ for each algorithm $A \in \A_{exp}$ and each $\alpha \in \P_{exp}$, and an expert $E(A)$ for each algorithm $A \in \A_{con}$.

Next, we employ Optimistic-Adapt-ML-Prod \citep{NIPS2016_405e2890} to track the best expert on the fly. In the beginning, we observe the regularization term $r(\cdot)$ (Step $3$). Then, at the $t$-th round, after receiving the predictions from all experts (Step $5$),  we are able to  compute the estimation $m_t^i$ and the weight $p_t^i$ of each expert $E^i$ (Steps $6$-$7$), and submit the weighted average $\x_t = \sum_{i=1}^{|\Ept|} p_t^i \x_t^i$ (Step $8$). Finally, we observe the loss function $f_t(\cdot)$ and $r(\cdot)$, and send the required information to each expert in $\Ept$ (Steps $9$-$10$).

The above process is similar to USC, and the distinction lies in  the update of  weight for each expert $E^i$. Specifically, the meta-algorithm employs the composite linearized loss in \eqref{eqn:com:linear:loss} to measure the performance of each expert. Because Optimistic-Adapt-ML-Prod requires $|\hat{R}_t^i - \hat{m}_t^i| \leq 2$, we normalize $\hat{R}_t^i$ and $\hat{m}_t^i$ in the following way
\begin{align}
     R_t^i = & \frac{1}{GD} \hat{R}_t^i \overset{\eqref{eqn:com:R:hat}}{=}  \frac{1}{GD} \left( \langle \nabla f_t(\x_t), \x_t- \x_t^i \rangle    +   \sum_{i=1}^{|\Ept|}  p_t^i r(\x_t^i)  -  r(\x_t^i) \right) \label{eqn:com:R} \\ 
     m_t^i = & \frac{1}{GD} \hat{m}_t^i \overset{\eqref{eqn:com:m:hat}}{=}  \frac{1}{GD}  \left( \sum_{i=1}^{|\Ept|}  p_t^i r(\x_t^i) - r(\x_t^i)\right). \label{eqn:com:m}
\end{align}
From \eqref{eqn:com:R} and \eqref{eqn:com:m}, we can verify that 
\[
    \left| R_t^i - m_t^i \right| = \frac{1}{GD}\left |  \langle \nabla f_t(\x_t), \x_t- \x_t^i \rangle \right| \leq \frac{1}{GD} \|\nabla f_t(\x_t)\|_2  \| \x_t- \x_t^i\|_2  \overset{\eqref{eqn:grad}, \eqref{eqn:domain}}{\leq} 2.
\]

\noindent
According to Optimistic-Adapt-ML-Prod \citep{NIPS2016_405e2890}, the weight of expert $E^i$ is computed by
\begin{equation} \label{eqn:com:pt:definition}
    p_t^i = \frac{\eta_{t-1}^i \wt_{t-1}^i }{\sum_{j=1}^{|\Ept|} \eta_{t-1}^j \wt_{t-1}^j}
\end{equation}
where $\wt_{t-1}^i = w_{t-1}^i \exp{\big(\eta_{t-1}^i m_{t}^i \big)}$, and  $\eta_{t-1}^i$ and $w_{t-1}^i$ are determined by 
\begin{align}
    &\eta_{t-1}^i =  \min\left\{ \frac{1}{4}, \sqrt{\frac{\ln |\Ept|}{1+\sum_{s=1}^{t-1} (R_s^i - m_s^i )^2 }} \right\}, \ t \geq 1, \\
    &w_{t-1}^i= \Big(w_{t-2}^i \exp \big(\eta_{t-2}^i R_{t-1}^i - \left(\eta_{t-2}^i (R_{t-1}^i - m_{t-1}^i )\right)^2 \big) \Big)^{\frac{\eta_{t-1}^i}{\eta_{t-2}^i}}, \ t \geq 2. \nonumber
\end{align}
In the beginning, we initialize  $w_0^i= 1/|\Ept|$.

Now, let's explain why the quantity $\sum_{i=1}^{|\Ept|}  p_t^i r(\x_t^i)$ of $m_t^i$ in \eqref{eqn:com:m} is computable before updating $p_t^i$. The key idea is to demonstrate that  $ \sum_{i=1}^{|\Ept|}  p_t^i r(\x_t^i) $  can be regarded as the fixed point of a continuous function. Given the value of $\gamma= \sum_{i=1}^{|\Ept|}  p_t^i r(\x_t^i)$, we have
\[
     m_t^i  \overset{\eqref{eqn:com:m}}{=}  \frac{1}{GD} \left( \sum_{i=1}^{|\Ept|}  p_t^i r(\x_t^i) - r(\x_t^i)\right) = \frac{1}{GD} \left( \gamma - r(\x_t^i)\right).
\]
Correspondingly,  $\wt_{t-1}^i$ and $p_t^i$ can be viewed as the function of $\gamma$, i.e., 
\begin{align*}
    \wt_{t-1}^i(\gamma)  = w_{t-1}^i \exp{\left(\frac{\eta_{t-1}^i}{GD} \left( \gamma - r(\x_t^i)\right) \right)} ~\text{and}~
    p_t^i(\gamma)  = \frac{\eta_{t-1}^i \wt_{t-1}^i(\gamma) }{\sum_{j=1}^{|\Ept|} \eta_{t-1}^j \wt_{t-1}^j(\gamma)}.
\end{align*}
Thus, the calculation of $\gamma$ reduces to finding the fixed point of $F(\gamma) = \sum_{i=1}^{|\Ept|}  p_t^i(\gamma) r(\x_t^i)$ satisfying $F(\gamma)=\gamma$. It can be verified that $F(\gamma)$ is continuous and bounded in $[0, C]$,  which implies that there must exist some fixed point $\gamma \in [0, C]$. To find the point $\gamma$, we can deploy the binary-search strategy, which only suffers  $1/T$ error in $\log T$ iterations and therefore, does not affect the regret bound \citep{NIPS2016_405e2890}.

\subsection{Strongly Convex Functions} 
\label{sec:str:com}
For the strongly convex   $f_t(\cdot)$ and convex   $r(\cdot)$,  we employ   algorithms below to build $\A_{\str}$:
\begin{compactitem}
    \item FOBOS for strongly convex functions (SC-FOBOS) \citep{Others:2009:Duchi};
    
    \item COMID for strongly convex functions (SC-COMID) \citep{COLT:2010:Composite};
    
    \item OCMD  for strongly convex functions (SC-OptCMD) \citep{TAC:2023:Scroccaro}.
\end{compactitem}
Note that \citet{COLT:2010:Composite} only provide the worst-case regret bounds of SC-COMID for strongly convex time-varying functions. In   Appendix~\ref{sec:app:COMID}, we further utilize the smoothness of $f_t(\cdot)$ and establish a \textit{new} (pseudo) small-loss bound of $O( \log \Lt_T )$ for smooth and strongly convex $f_t(\cdot)$, where
\begin{equation}\label{eqn:L:til}
    \Lt_T =  \sum_{t=1}^T f_t(\tilde{\x}_t) 
\end{equation}
is the cumulative loss of SC-COMID involving $f_t(\cdot)$.

\noindent
\textbf{Remark:}
We emphasize that $O( \log \Lt_T )$ is not the standard  small-loss bound, since  $\Lt_T$ depends on not only  the time-varying function  $f_t(\cdot)$ but also the prediction $\tilde{\x}_t$ chosen  by the algorithm itself. Due to the additional regularization term $r(\cdot)$ in \eqref{eqn:com:regret}, it remains unclear how to establish a standard  small-loss bound that solely  depends on $f_t(\cdot)$.

Following the same configuration as in Section~\ref{sec:strong:convex}, we assume the modulus of strong convexity $\lambda \in [1/T, 1]$, and construct $\P_{\str}$ in the same way as \eqref{eqn:discrete:lambda}, according to which there must exist a $\lbah \in \P_{\str}$ such that $\lbah \leq \lambda \leq 2 \lbah$ for any $\lambda \in [1/T,1]$. We denote by $R(A,\lbah)$ the regret bound of expert $E(A,\lbah)$ in Algorithm~\ref{alg:2}. Since $\lbah \leq \lambda \leq 2 \lbah$, $R(A,\lbah)$ is both valid and tight. We have the following theorem for strongly convex functions.

\begin{theorem} \label{thm:com:USC:strong} 
    Under Assumptions~\ref{ass:1},~\ref{ass:2},~\ref{ass:4}~and~\ref{ass:5}, if the time-varying function $f_t(\cdot)$ is $\lambda$-strongly convex with $\lambda \in [1/T, 1]$, Algorithm~\ref{alg:2} satisfies
    \[
        \begin{split}
            & \sum_{t=1}^T \left [f_t(\x_t) + r(\x_t)\right] - \min_{\x \in \X} \sum_{t=1}^T \left [ f_t(\x) + r(\x) \right]  \\
            \leq &\min_{A \in \A_{\str}}  R(A,\lbah)+ GD \left( \Xi  + \frac{\Psi}{\sqrt{\ln |\Ept|}} \right)  + \frac{\Psi^2 G^2}{2 \lambda \ln |\Ept|} 
            =  \min_{A \in \A_{\str}}  R(A,\lbah)+  O \left( \frac{\log \log T}{\lambda} \right)
         \end{split}
    \]
    where $\lbah \in \P_{\str}$, $\lbah \leq \lambda \leq 2 \lbah$, and
    \begin{align} 
            \Psi = & \ln |\Ept| + \ln \left( 1 + \frac{|\Ept|}{ e} \big(1 + \ln (T+1)\big)\right)  \overset{\eqref{eqn:number:experts}}{=}  O(\log \log T), \label{eqn:psi-com:definiton}  \\ 
            \Xi  = & \frac{1}{4} \Psi + 2 \sqrt{\ln |\Ept|} + 16 \ln |\Ept|   \overset{\eqref{eqn:number:experts}}{=}  O(\log \log T). \label{eqn:xi-com:definiton}
    \end{align}

\end{theorem}
\textbf{Remark:} The above theorem indicates that Algorithm~\ref{alg:2} preserves the regret bound of \emph{any} expert designed for strongly convex functions, up to a negligible double logarithmic factor. Specifically, when the time-varying function $f_t(\cdot)$ is smooth, we can achieve the following problem-dependent bound, by using SC-COMID \citep{COLT:2010:Composite} and SC-OptCMD \citep{TAC:2023:Scroccaro} as experts.

\begin{corollary} \label{cor:com:str}  
    Under Assumptions~\ref{ass:1},~\ref{ass:2},~\ref{ass:3},~\ref{ass:4}~and~\ref{ass:5}, if the time-varying function $f_t(\cdot)$ is $\lambda$-strongly convex with $\lambda \in [1/T, 1]$, and  $\textrm{SC-COMID}, \textrm{SC-OptCMD} \in \A_{\str}$,  we have
    \[
    \sum_{t=1}^T \left [f_t(\x_t) + r(\x_t)\right] - \min_{\x \in \X} \sum_{t=1}^T \left [ f_t(\x) + r(\x) \right] = O \left( \frac{1}{\lambda} \Big( \min\{\log \Lt_T, \log V_T\}  + \log\log T \Big) \right)
    \]
    where $\Lt_T$  and $V_T$ are defined in \eqref{eqn:L:til} and  \eqref{eqn:grad:var} respectively.
\end{corollary}

\subsection{Exp-concave Functions}
For the exp-concave    $f_t(\cdot)$ and convex   $r(\cdot)$,  we employ      algorithms below to build $\A_{\exp}$:
\begin{compactitem}
    \item ProxONS \citep{Others:2023:Yang};
    
    \item OCMD  for exp-concave functions (Exp-OptCMD) \citep{TAC:2023:Scroccaro}.
\end{compactitem}
Note that \citet{TAC:2023:Scroccaro} merely  investigate OCMD for general convex and strongly convex time-varying functions, under the smoothness condition of $f_t(\cdot)$. In Appendix~\ref{sec:app:optcmd}, we further extend OCMD to exp-concave time-varying functions, and  establish \textit{new}  problem-dependent bounds of $O(\log V_T)$ and $O(\log \Lt_T)$.

Similar to Section~\ref{sec:exp:concave}, we  assume the modulus of exponential concavity $\alpha$ lies in $[1/T,1]$, and use the same geometric series to construct $\P_{\expc}$ as \eqref{eqn:discrete:alpha}. Then, for any $\alpha \in [1/T,1]$, there must exist an $\laha \in \P_{\expc}$ such that $\laha \leq \alpha \leq 2 \laha$.  We denote by $R(A,\laha)$ the regret bound of expert $E(A,\laha)$ in Algorithm~\ref{alg:2}, and have the following guarantee for exp-concave functions.

\begin{theorem} \label{thm:com:USC:exp} 
    Under Assumptions~\ref{ass:1},~\ref{ass:2},~\ref{ass:4}~and~\ref{ass:5}, if the time-varying function $f_t(\cdot)$ is $\alpha$-exp-concave with $\alpha \in [1/T, 1]$, Algorithm~\ref{alg:2} satisfies
    \[
        \begin{split}
            & \sum_{t=1}^T \left [f_t(\x_t) + r(\x_t)\right] - \min_{\x \in \X} \sum_{t=1}^T \left [ f_t(\x) + r(\x) \right] \\
            \leq & \min_{A \in \A_{\expc}}  R(A,\laha)+   GD \left( \Xi  + \frac{\Psi}{\sqrt{\ln |\Ept|}} \right)  + \frac{\Psi^2 }{2 \beta \ln |\Ept|} 
            =  \min_{A \in \A_{\expc}}  R(A,\laha)+  O \left( \frac{\log \log T}{\alpha} \right)
         \end{split}
    \]
    where $\laha \in \P_{\expc}$, $\laha \leq \alpha \leq 2 \laha$, $\beta = \frac{1}{2} \min\{\frac{1}{4GD}, \alpha\}$, and $\Psi$ and $\Xi$ are defined in \eqref{eqn:psi-com:definiton} and \eqref{eqn:xi-com:definiton}, respectively.
\end{theorem}
\textbf{Remark:} Similar to the strongly convex case, Algorithm~\ref{alg:2} also inherits  the regret bound of any expert designed for exp-concave functions. By incorporating Exp-OptCMD into $\A_{\exp}$, Algorithm~\ref{alg:2} ensures the following problem-dependent bound.

\begin{corollary} \label{cor:com:exp}  
    Under Assumptions~\ref{ass:1},~\ref{ass:2},~\ref{ass:3},~\ref{ass:4}~and~\ref{ass:5}, if the time-varying function $f_t(\cdot)$ is $\alpha$-exp-concave with $\alpha \in [1/T, 1]$, we have
    \[
        \sum_{t=1}^T \left [f_t(\x_t) + r(\x_t)\right] - \min_{\x \in \X} \sum_{t=1}^T \left [ f_t(\x) + r(\x) \right] = O \left( \frac{1}{\alpha} \Big( d\min\{\log \Lt_T, \log V_T\}  + \log\log T \Big) \right)
    \]
    where $\Lt_T$  and $V_T$ are defined in \eqref{eqn:L:til} and  \eqref{eqn:grad:var} respectively, provided $\textrm{Exp-OptCMD} \in \A_{\str}$.
\end{corollary}

\subsection{General Convex Functions} 
For the general convex   $f_t(\cdot)$ and convex   $r(\cdot)$,  we employ     algorithms below to build $\A_{\con}$:
\begin{compactitem}
    \item FOBOS \citep{Others:2009:Duchi};
    
    \item COMID \citep{COLT:2010:Composite};
    
    \item CAO-FTRL \citep{AISTATS:2016:Mohri};
    
    \item Composite-objective AO-FTRL  \citep{Others:2020:Joulani};
    
    \item OCMD  \citep{TAC:2023:Scroccaro}.
\end{compactitem}
During the analysis,  we  develop a \textit{new} (pseudo) small-loss bound of $O(\sqrt{\Lt_T} )$ for COMID when the time-varying function $f_t(\cdot)$ is smooth and general convex. More details can be found in   Appendix~\ref{sec:app:COMID}.

Let $R(A)$ be the regret bound of expert $E(A)$ in $\A_{\con}$. The theoretical guarantee of Algorithm~\ref{alg:2} for general convex functions is stated below.

\begin{theorem} \label{thm:com:USC:con} 
    Under Assumptions~\ref{ass:1},~\ref{ass:2},~\ref{ass:4}~and~\ref{ass:5}, for any sequence of convex time-varying functions, Algorithm~\ref{alg:2} satisfies
    \[
        \begin{split}
            &\sum_{t=1}^T \left [f_t(\x_t) + r(\x_t)\right] - \min_{\x \in \X} \sum_{t=1}^T \left [ f_t(\x) + r(\x) \right] \\
            \leq & \min_{A \in \A_{\con}}  R(A)+    \Xi GD   +   \frac{\Psi D}{\sqrt{\ln |\Ept|}} \sqrt{G^2  + \sum_{t=1}^T \| \nabla f_t(\x_t) \|_2^2} 
            =  \min_{A \in \A_{\con}}  R(A)+  O \left( \sqrt{T \log\log T} \right)
         \end{split}
    \]
    where   $\Psi$ and $\Xi$ is defined in \eqref{eqn:psi-com:definiton} and \eqref{eqn:xi-com:definiton}, respectively.
\end{theorem}
\textbf{Remark}:
The above bound remains weaker than those in Theorems~\ref{thm:com:USC:strong}~and~\ref{thm:com:USC:exp}, due to the uneliminated component $O(\sqrt{\sum_{t=1}^T \| \nabla f_t(\x_t) \|_2^2})$. 
Under the smoothness of $f_t(\cdot)$,   by employing $\textrm{COMID}$ as an expert in $\A_{\con}$,  Theorem~\ref{thm:com:USC:con} implies the following problem-dependent regret bound.

\begin{corollary} \label{cor:com:con}  
    Under Assumptions~\ref{ass:1},~\ref{ass:2},~\ref{ass:3},~\ref{ass:4}~and~\ref{ass:5}, for any sequence of convex time-varying functions, we have
    \[
        \sum_{t=1}^T \left [f_t(\x_t) + r(\x_t)\right] - \min_{\x \in \X} \sum_{t=1}^T \left [ f_t(\x) + r(\x) \right] = O\left( \sqrt{\Lt_{T}^{meta} \log\log T} + \sqrt{\Lt_{T}} \right)
    \]
    where $\Lt_{T}^{meta}$ denotes the   cumulative loss in \eqref{eqn:L:til} of the meta-algorithm, and $\Lt_T$ denotes that  of COMID, provided $\textrm{COMID} \in \A_{\con}$.
\end{corollary}

\section{Analysis} \label{sec:analysis}
In this section, we present the analysis of all theorems.

\subsection{Proof of Theorem~\ref{thm:USC:strong}}
We first analyze the meta-regret of our strategy. According to the theoretical guarantee of Adapt-ML-Prod \citep[Corollary 4]{pmlr-v35-gaillard14}, we have
\[
    \sum_{t=1}^T \ell_t - \sum_{t=1}^T \ell_t^i \leq \frac{\Gamma}{\sqrt{\ln |\Ept|}} \sqrt{1 + \sum_{t=1}^T (\ell_t -  \ell_t^i)^2} + 2 \Gamma
\]
for all expert $E^i \in \Ept$, where $\Gamma$ is given in (\ref{eqn:beta:definiton}). Combining with the definitions of $\ell_t^i$ and $\ell_t$ in (\ref{eqn:linear:loss}) and (\ref{eqn:meta:loss}), we arrive at
\begin{equation} \label{eqn:meta:regret:1}
    \begin{split}
         \sum_{t=1}^T  \langle \nabla f_t(\x_t), \x_t- \x_t^i \rangle 
         \leq&    4 \Gamma GD   +   \frac{\Gamma}{\sqrt{\ln |\Ept|}} \sqrt{4G^2D^2 + \sum_{t=1}^T \langle \nabla f_t(\x_t), \x_t- \x_t^i \rangle^2} \\
         \leq &  2\Gamma GD \left(2 + \frac{1}{\sqrt{\ln |\Ept|}} \right)  +   \frac{\Gamma}{\sqrt{\ln |\Ept|}} \sqrt{\sum_{t=1}^T \langle \nabla f_t(\x_t), \x_t- \x_t^i \rangle^2}
    \end{split}
\end{equation}
where the last step follows from the basic inequality $\sqrt{a+b} \leq \sqrt{a} + \sqrt{b}$,  $\forall a, b \geq 0$.

To utilize the property of strong convexity in \eqref{eqn:strong:convex:def}, we proceed in the following way:
\begin{equation} \label{eqn:meta:regret:2}
    \begin{split}
        & \sum_{t=1}^T  \langle \nabla f_t(\x_t), \x_t- \x_t^i \rangle  \\
        \leq & 2\Gamma GD \left(2 + \frac{1}{\sqrt{\ln |\Ept|}} \right) + \frac{\Gamma^2 G^2}{2 \lambda \ln |\Ept|}  +  \frac{\lambda}{2 G^2} \sum_{t=1}^T \langle \nabla f_t(\x_t), \x_t- \x_t^i \rangle^2 \\
        \leq & 2\Gamma GD \left(2 + \frac{1}{\sqrt{\ln |\Ept|}} \right) + \frac{\Gamma^2 G^2}{2 \lambda \ln |\Ept|}  +  \frac{\lambda}{2 G^2} \sum_{t=1}^T \|\nabla f_t(\x_t)\|_2^2  \|\x_t- \x_t^i \|_2^2 \\
        \overset{\eqref{eqn:grad}}{\leq} & 2\Gamma GD \left(2 + \frac{1}{\sqrt{\ln |\Ept|}} \right) + \frac{\Gamma^2 G^2}{2 \lambda \ln |\Ept|} +  \frac{\lambda}{2 } \sum_{t=1}^T  \|\x_t- \x_t^i \|_2^2
    \end{split}
\end{equation}
where the first step follows from the basic inequality $2 \sqrt{ab} \leq a + b$, $\forall a, b \geq 0$. According to Definition~\ref{def:strong}, the meta-regret in terms of $f_t(\cdot)$ is given by
\begin{equation} \label{eqn:meta:regret:3}
    \begin{split}
        \sum_{t=1}^T    f_t(\x_t)- \sum_{t=1}^T f_t(\x_t^i)   \overset{\eqref{eqn:strong:convex:def}}{\leq} &  \sum_{t=1}^T  \left(\langle \nabla f_t(\x_t), \x_t -\x_t^i   \rangle  - \frac{\lambda}{2} \| \x_t -\x_t^i\|_2^2 \right) \\
        \overset{\eqref{eqn:meta:regret:2}}{\leq} &  2\Gamma GD \left(2 + \frac{1}{\sqrt{\ln |\Ept|}} \right) + \frac{\Gamma^2 G^2}{2 \lambda \ln |\Ept|}.
    \end{split}
\end{equation}

Next, we study the expert-regret. Let $E^i$ be the expert $E(A,\lbah)$ where $A \in \A_{\str}$, $\lbah \in \P_{\str}$, and $\lbah \leq \lambda \leq 2 \lbah$. Since $\lambda$-strongly convex functions are also $\lbah$-strongly convex, expert $E(A,\lbah)$ makes a right assumption, and the following inequality is true
\begin{equation} \label{eqn:expert:regret}
    \sum_{t=1}^T f_t(\x_t^i) - \min_{\x \in \X} \sum_{t=1}^T f_t(\x)  \leq R(A,\lbah).
\end{equation}
Combining (\ref{eqn:meta:regret:3}) and (\ref{eqn:expert:regret}), we have
\begin{equation}  \label{eqn:overall:regret}
    \begin{split}
        \sum_{t=1}^T    f_t(\x_t)- \min_{\x \in \X} \sum_{t=1}^T f_t(\x)   \leq R(A,\lbah) +   2\Gamma GD \left(2 + \frac{1}{\sqrt{\ln |\Ept|}} \right) + \frac{\Gamma^2 G^2}{2 \lambda \ln |\Ept|}.
    \end{split}
\end{equation}
We complete the proof by noticing that (\ref{eqn:overall:regret}) holds for any $A \in \A_{\str}$.

\subsection{Proof of Corollary~\ref{cor:1}}
From the theoretical guarantee of OEGD for strongly convex and smooth functions in Theorem~\ref{thm:OEGD:Strong}, we have
\begin{equation} \label{eqn:OEGD:strong:smooth}
    R(\textrm{OEGD},\lbah)= O\left(\frac{\log V_T}{\lbah} \right) \overset{\lambda \leq 2 \lbah}{=} O\left(\frac{\log V_T}{\lambda} \right).
\end{equation}
Similarly, from the regret bound of S$^2$OGD \citep[Theorem 1]{AAAI:2020:Wang}, we have
\begin{equation} \label{eqn:S2OGD:strong:smooth}
    R(\textrm{S}^2\textrm{OGD},\lbah)= O\left(\frac{\log L_T^*}{\lbah} \right) \overset{\lambda \leq 2 \lbah}{=} O\left(\frac{\log L_T^*}{\lambda} \right).
\end{equation}
We obtain the corollary by substituting (\ref{eqn:OEGD:strong:smooth}) and (\ref{eqn:S2OGD:strong:smooth}) into Theorem~\ref{thm:USC:strong}.

\subsection{Proof of Theorem~\ref{thm:USC:exp}}
The analysis is similar to that of Theorem~\ref{thm:USC:strong}. To make use of the property of exponential concavity in (\ref{eqn:exp:concave:prop}), we change (\ref{eqn:meta:regret:2}) as follows:
\begin{equation} \label{eqn:meta:regret:4}
    \begin{split}
         & \sum_{t=1}^T  \langle \nabla f_t(\x_t), \x_t- \x_t^i \rangle  \\
         \leq & 2\Gamma GD \left(2 + \frac{1}{\sqrt{\ln |\Ept|}} \right) + \frac{\Gamma^2 }{2 \beta \ln |\Ept|}   +  \frac{\beta}{2 } \sum_{t=1}^T \langle \nabla f_t(\x_t), \x_t- \x_t^i \rangle^2.
    \end{split}
\end{equation}

According to Lemma~\ref{lem:exp}, the meta-regret in terms of $f_t(\cdot)$ can be bounded by
\[
    \begin{split}
        \sum_{t=1}^T    f_t(\x_t)- \sum_{t=1}^T f_t(\x_t^i)  
        \overset{\eqref{eqn:exp:concave:prop}}{\leq} &  \sum_{t=1}^T \left(\langle \nabla f_t(\x_t), \x_t -\x_t^i   \rangle  - \frac{\beta}{2}   \langle \nabla f_t(\x_t),  \x_t -\x_t^i \rangle^2 \right) \\
        \overset{\eqref{eqn:meta:regret:4}}{\leq} &  2\Gamma GD \left(2 + \frac{1}{\sqrt{\ln |\Ept|}} \right) + \frac{\Gamma^2 }{2 \beta \ln |\Ept|}.
    \end{split}
\]
The rest of the proof is identical to that of Theorem~\ref{thm:USC:strong}.

\subsection{Proof of Corollary~\ref{cor:2}}
From the theoretical guarantee of ONS for exp-concave and smooth functions \citep[Theorem 1]{Beyond:Logarithmic}, we have
\begin{equation} \label{eqn:ONS:exp:smooth}
    R(\textrm{ONS},\laha)= O\left(\frac{d \log L_T^*}{\laha} \right) \overset{\alpha \leq 2 \laha}{=} O\left(\frac{d \log L_T^*}{\alpha} \right).
\end{equation}
Similarly, from the regret bound of OEGD \citep[Theorem 15]{Gradual:COLT:12}, we have
\begin{equation} \label{eqn:OEGD:exp:smooth}
    R(\textrm{OEGD},\laha)= O\left(\frac{d \log V_T}{\laha} \right) \overset{\alpha \leq 2 \laha}{=} O\left(\frac{d \log V_T}{\alpha} \right).
\end{equation}
We obtain the corollary by substituting (\ref{eqn:ONS:exp:smooth}) and (\ref{eqn:OEGD:exp:smooth}) into Theorem~\ref{thm:USC:exp}.

\subsection{Proof of Theorem~\ref{thm:USC:convex}}
From the first-order condition of convex functions, we bound the meta-regret by
\[
    \begin{split}
        &\sum_{t=1}^T    f_t(\x_t)- f_t(\x_t^i)  \leq   \sum_{t=1}^T \langle \nabla f_t(\x_t), \x_t -\x_t^i   \rangle   \\
        \overset{\eqref{eqn:meta:regret:1}}{\leq} &  4 \Gamma GD   +   \frac{\Gamma}{\sqrt{\ln |\Ept|}} \sqrt{4G^2D^2 + \sum_{t=1}^T \langle \nabla f_t(\x_t), \x_t- \x_t^i \rangle^2} \\
        \leq &  4 \Gamma GD   +   \frac{\Gamma}{\sqrt{\ln |\Ept|}} \sqrt{4G^2D^2 + \sum_{t=1}^T \| \nabla f_t(\x_t)\|_2^2 \| \x_t- \x_t^i \|_2^2} \\
        \overset{\eqref{eqn:domain}}{\leq} & 4 \Gamma GD   +   \frac{\Gamma D}{\sqrt{\ln |\Ept|}} \sqrt{4G^2 + \sum_{t=1}^T  \| \nabla f_t(\x_t) \|_2^2}.
    \end{split}
\]
We complete the proof by combining the above inequality with that of the expert-regret:
\[
    \sum_{t=1}^T f_t(\x_t^i) - \min_{\x \in \X} \sum_{t=1}^T f_t(\x)  \leq R(A), \ \forall A \in \A_{\con}.
\]

\subsection{Proof of Corollary~\ref{cor:3}}
We need the self-bounding property of smooth functions \citep[Lemma 3.1]{NIPS2010_Smooth}.
\begin{lemma} \label{lem:smooth} 
    For a nonnegative and $H$-smooth function $f: \X \mapsto \R$, we have
    \begin{equation} \label{eqn:smoothness:property}
        \| \nabla f(\x)\| \leq \sqrt{4 H f(\x)}, \ \forall \x \in \X.
    \end{equation}
\end{lemma}

Combining Lemma~\ref{lem:smooth} and Theorem~\ref{thm:USC:convex}, we have
\begin{equation} \label{eqn:convex:smooth:regret:1}
    \sum_{t=1}^T f_t(\x_t) - \min_{\x \in \X} \sum_{t=1}^T f_t(\x) \leq \min_{A \in \A_{\con}}  R(A)      +4 \Gamma GD   +   \frac{\Gamma D}{\sqrt{\ln |\Ept|}} \sqrt{4G^2 + 4H \sum_{t=1}^T   f_t(\x_t)}.
\end{equation}
From the theoretical guarantee of SOGD for convex and smooth functions \citep[Theorem 2]{Adaptive:Regret:Smooth:ICML}, we have
\begin{equation} \label{eqn:SOGD:convex:smooth}
    R(\textrm{SOGD})=  8 H D^2 +D\sqrt{2 \delta + 8 H L_T^*}
\end{equation}
where $\delta >0$ can be any small constant. Substituting (\ref{eqn:SOGD:convex:smooth}) into (\ref{eqn:convex:smooth:regret:1}), we obtain
\begin{equation} \label{eqn:convex:smooth:regret:2}
    \sum_{t=1}^T f_t(\x_t) - L_T^* \leq 8 H D^2 +D\sqrt{2 \delta + 8 H L_T^*}    +4 \Gamma GD   +   \frac{\Gamma D}{\sqrt{\ln |\Ept|}} \sqrt{4G^2 + 4H \sum_{t=1}^T   f_t(\x_t)}.
\end{equation}

To simplify the above inequality, we use the following lemma \citep[Lemma 19]{Shai:thesis}.
\begin{lemma} \label{lem:tool} 
    Let $x, b,c \in \R_+$. Then,
    \[
        x  -c \leq b \sqrt{x} \Rightarrow  x - c \leq b^2 + b \sqrt{c}.
    \]
\end{lemma}
From \eqref{eqn:convex:smooth:regret:2}, we have
\[
    \begin{split}
        &\left(\frac{G^2}{H} + \sum_{t=1}^T f_t(\x_t)  \right)- \left( L_T^* +D\sqrt{2 \delta + 8 H L_T^*} + 4 \Gamma GD  +8 H D^2 +\frac{G^2}{H}    \right)\\
        \leq & \frac{\Gamma D\sqrt{4H}}{\sqrt{\ln |\Ept|}} \sqrt{\frac{G^2}{H} +  \sum_{t=1}^T   f_t(\x_t)} .
    \end{split}
\]
Lemma~\ref{lem:tool} implies
\[
    \begin{split}
        &\left(\frac{G^2}{H} + \sum_{t=1}^T f_t(\x_t)  \right)- \left( L_T^* +D\sqrt{2 \delta + 8 H L_T^*} + 4 \Gamma GD  +8 H D^2 +\frac{G^2}{H}    \right) \\
        \leq & \frac{4 \Gamma^2 D^2 H}{\ln |\Ept|} +  \frac{\Gamma D\sqrt{4H}}{\sqrt{\ln |\Ept|}} \sqrt{L_T^* +D\sqrt{2 \delta + 8 H L_T^*} + 4 \Gamma GD  +8 H D^2 +\frac{G^2}{H}}.
    \end{split}
\]
Thus,
\[
    \begin{split}
        & \sum_{t=1}^T f_t(\x_t) - \min_{\x \in \X} \sum_{t=1}^T f_t(\x) = \sum_{t=1}^T f_t(\x_t)  - L_T^*  \\
        \leq &  \frac{\Gamma D\sqrt{4H}}{\sqrt{\ln |\Ept|}} \sqrt{L_T^* +D\sqrt{2 \delta + 8 H L_T^*} + 4 \Gamma GD  +8 H D^2 +\frac{G^2}{H}} +D\sqrt{2 \delta + 8 H L_T^*} \\
        & + 4 \Gamma GD  +8 H D^2 + \frac{4 \Gamma^2 D^2 H}{\ln |\Ept|} \\
        =& O\left(\sqrt{L_T^* \log \log T}\right) .
    \end{split}
\]

\subsection{Proof of Theorem~\ref{thm:com:USC:strong}}
Similar to the analysis of Theorem~\ref{thm:USC:strong}, we first bound the meta-regret of Algorithm~\ref{alg:2}. According to the theoretical guarantee of Optimistic-Adapt-ML-Prod \citep[Proof of Theorem 3.4]{NIPS2016_405e2890}, we have
\begin{equation}    \label{eqn:com:meta:regret:0}
    \sum_{t=1}^T R_t^i \leq \frac{\Psi}{\sqrt{\ln |\Ept|}} \sqrt{1 + \sum_{t=1}^T (R_t^i -  m_t^i)^2} + \Xi 
\end{equation}
for all expert $E^i \in \Ept$, where $\Psi$ and $\Xi $ are given in \eqref{eqn:psi-com:definiton} and \eqref{eqn:xi-com:definiton}, respectively. Then, we substitute  \eqref{eqn:com:R} and \eqref{eqn:com:m} into \eqref{eqn:com:meta:regret:0}, and obtain
\begin{equation} \label{eqn:com:meta:regret:1}
    \begin{split}
        &  \sum_{t=1}^T  \langle \nabla f_t(\x_t), \x_t- \x_t^i \rangle + \sum_{t=1}^T \sum_{i = 1}^{|\Ept|} p_t^i r(\x_t^i) -   \sum_{t=1}^T  r(\x_t^i)\\
        \leq  &   \Xi GD   +   \frac{\Psi}{\sqrt{\ln |\Ept|}} \sqrt{G^2D^2 + \sum_{t=1}^T \langle \nabla f_t(\x_t), \x_t- \x_t^i \rangle^2} \\
        \leq &    GD \left( \Xi  + \frac{\Psi}{\sqrt{\ln |\Ept|}} \right)  +   \frac{\Psi}{\sqrt{\ln |\Ept|}} \sqrt{\sum_{t=1}^T \langle \nabla f_t(\x_t), \x_t- \x_t^i \rangle^2}
    \end{split}
\end{equation}
where the last step follows from the basic inequality $\sqrt{a+b} \leq \sqrt{a} + \sqrt{b}$, $\forall a, b \geq 0$.

To utilize the property of strong convexity in \eqref{eqn:strong:convex:def}, we simplify \eqref{eqn:com:meta:regret:1} into the following form:
\begin{equation} \label{eqn:com:meta:regret:2}
    \begin{split}
        & \sum_{t=1}^T  \langle \nabla f_t(\x_t), \x_t- \x_t^i \rangle  + \sum_{t=1}^T \sum_{i = 1}^{|\Ept|} p_t^i r(\x_t^i) -   \sum_{t=1}^T  r(\x_t^i)  \\
         \leq & GD \left( \Xi  + \frac{\Psi}{\sqrt{\ln |\Ept|}} \right)  + \frac{\Psi^2 G^2}{2 \lambda \ln |\Ept|}  +  \frac{\lambda}{2 G^2} \sum_{t=1}^T \langle \nabla f_t(\x_t), \x_t- \x_t^i \rangle^2 \\
         \leq & GD \left( \Xi  + \frac{\Psi}{\sqrt{\ln |\Ept|}} \right)  + \frac{\Psi^2 G^2}{2 \lambda \ln |\Ept|}  +  \frac{\lambda}{2 G^2} \sum_{t=1}^T \|\nabla f_t(\x_t)\|_2^2  \|\x_t- \x_t^i \|_2^2 \\
         \overset{\eqref{eqn:grad}}{\leq} & GD \left( \Xi  + \frac{\Psi}{\sqrt{\ln |\Ept|}} \right)  + \frac{\Psi^2 G^2}{2 \lambda \ln |\Ept|} +  \frac{\lambda}{2 } \sum_{t=1}^T  \|\x_t- \x_t^i \|_2^2
    \end{split}
\end{equation}
where the first step follows from the basic inequality $2 \sqrt{ab} \leq a + b$, $\forall a, b \geq 0$. According to \eqref{eqn:strong:convex:def} and Jensen's inequality, i.e., $ r(\x_t) =   r( \sum_{i=1}^{|\Ept|} p_t^i \x_t^i)  \leq  \sum_{i=1}^{|\Ept|}  p_t^i r(\x_t^i) $, the meta-regret  is upper bounded by
\begin{equation}       \label{eqn:com:meta:regret:2.5}
    \begin{split}
        & \sum_{t=1}^T  \left [ f_t(\x_t) + r(\x_t) \right]- \sum_{t=1}^T \left [ f_t(\x_t^i) + r(\x_t^i) \right ]  \\ \overset{\eqref{eqn:strong:convex:def}}{\leq} &  \sum_{t=1}^T  \left(\langle \nabla f_t(\x_t), \x_t -\x_t^i   \rangle  - \frac{\lambda}{2} \| \x_t -\x_t^i\|_2^2 \right) + \sum_{t=1}^T \sum_{i = 1}^{|\Ept|} p_t^i r(\x_t^i) -   \sum_{t=1}^T  r(\x_t^i).
    \end{split}
\end{equation}
Substituting \eqref{eqn:com:meta:regret:2} into \eqref{eqn:com:meta:regret:2.5}, we arrive at
\begin{equation} \label{eqn:com:meta:regret:3}
    \begin{split}
        & \sum_{t=1}^T  \left [ f_t(\x_t) + r(\x_t) \right]- \sum_{t=1}^T \left [ f_t(\x_t^i) + r(\x_t^i) \right ] 
         \leq   GD \left( \Xi  + \frac{\Psi}{\sqrt{\ln |\Ept|}} \right)  + \frac{\Psi^2 G^2}{2 \lambda \ln |\Ept|}.
    \end{split}
\end{equation}

\noindent
We complete the proof by combining \eqref{eqn:com:meta:regret:3} with the following  expert-regret 
\begin{equation} \label{eqn:com:expert:regret}
    \sum_{t=1}^T \left [f_t(\x_t^i) + r(\x_t^i)\right] - \min_{\x \in \X} \sum_{t=1}^T \left [ f_t(\x) + r(\x) \right]   \leq R(A,\lbah), \ \forall A \in \A_{\str},  \lbah \in \P_{\str}.
\end{equation}

\subsection{Proof of Corollary~\ref{cor:com:str}}
From the regret bound of SC-COMID for strongly convex and smooth functions in Theorem~\ref{thm:COMID:str}, we have
\begin{equation} \label{eqn:com:str:1}
    \begin{split}
        R(\textrm{SC-COMID}, \lbah) =  O\left(\frac{\log \Lt_T}{\lbah} \right) \overset{\lambda \leq 2 \lbah}{=} O\left(\frac{\log \Lt_T}{\lambda} \right).
    \end{split}
\end{equation}
Similarly, from the regret bound of SC-OptCMD \citep[Theorem 2.9]{TAC:2023:Scroccaro}, we have
\begin{equation} \label{eqn:com:str:2}
    R(\textrm{SC-OptCMD},\lbah)= O\left(\frac{\log V_T}{\lbah} \right) \overset{\lambda \leq 2 \lbah}{=} O\left(\frac{\log V_T}{\lambda} \right).
\end{equation}
We complete the proof by substituting \eqref{eqn:com:str:1} and \eqref{eqn:com:str:2} into Theorem~\ref{thm:com:USC:strong}.

\subsection{Proof of Theorem~\ref{thm:com:USC:exp}}
Similar to   \eqref{eqn:com:meta:regret:2}, we can obtain the following bound for the exp-concave case 
\begin{equation}    \label{eqn:com:exp:meta:regret:1}
    \begin{split}
         &\sum_{t=1}^T  \langle \nabla f_t(\x_t), \x_t- \x_t^i \rangle  + \sum_{t=1}^T \sum_{i = 1}^{|\Ept|} p_t^i r(\x_t^i) -   \sum_{t=1}^T  r(\x_t^i)  \\
         \leq & GD \left( \Xi  + \frac{\Psi}{\sqrt{\ln |\Ept|}} \right)  + \frac{\Psi^2 }{2 \beta \ln |\Ept|} +  \frac{\beta}{2 } \sum_{t=1}^T \langle \nabla f_t(\x_t), \x_t- \x_t^i \rangle^2.
    \end{split}
\end{equation}
Then, substituting \eqref{eqn:com:exp:meta:regret:1} into \eqref{eqn:com:meta-regret}, we  obtain 
\begin{equation}     \label{eqn:com:exp:meta:regret:2}
    \begin{split}
        & \sum_{t=1}^T  \left [ f_t(\x_t) + r(\x_t) \right]- \sum_{t=1}^T \left [ f_t(\x_t^i) + r(\x_t^i) \right ] 
         \leq  GD \left( \Xi  + \frac{\Psi}{\sqrt{\ln |\Ept|}} \right)  + \frac{\Psi^2 }{2 \beta \ln |\Ept|}.
    \end{split}
\end{equation}
The rest of the proof is identical to that of Theorem~\ref{thm:com:USC:strong}.

\subsection{Proof of Corollary~\ref{cor:com:exp}}
From the regret bounds of Exp-OptCMD  in Theorems~\ref{thm:OCMD:exp:VT}~and~\ref{thm:OCMD:exp:LT}, we have
\begin{equation} \label{eqn:com:exp:1}
    \begin{split}
        R(\textrm{Exp-OptCMD},\laha)= O\left(\frac{d }{\laha} \min \{\log V_T, \log \Lt_T\}\right) \overset{\alpha \leq 2 \laha}{=} O\left(\frac{d}{\alpha} \min \{\log V_T, \log \Lt_T\} \right).
    \end{split}
\end{equation}
We complete the proof by substituting \eqref{eqn:com:exp:1} into Theorem~\ref{thm:com:USC:exp}.

\subsection{Proof of Theorem~\ref{thm:com:USC:con}}
We follow the similar analysis in \eqref{eqn:com:meta:regret:1} to handle the general convex case:
\begin{equation} \label{eqn:com:con:meta:regret:1}
    \begin{split}
        &  \sum_{t=1}^T  \langle \nabla f_t(\x_t), \x_t- \x_t^i \rangle + \sum_{t=1}^T \sum_{i = 1}^{|\Ept|} p_t^i r(\x_t^i) -   \sum_{t=1}^T  r(\x_t^i)\\
         \leq  &   \Xi GD   +   \frac{\Psi}{\sqrt{\ln |\Ept|}} \sqrt{G^2D^2 + \sum_{t=1}^T \langle \nabla f_t(\x_t), \x_t- \x_t^i \rangle^2} \\
         \overset{\eqref{eqn:domain}}{\leq}   &   \Xi GD   +   \frac{\Psi D}{\sqrt{\ln |\Ept|}} \sqrt{G^2  + \sum_{t=1}^T \| \nabla f_t(\x_t) \|_2^2}.
    \end{split}
\end{equation}
Then, we complete the proof by combining \eqref{eqn:com:con:meta:regret:1} with the following expert regret bound 
\begin{equation} 
    \sum_{t=1}^T \left [f_t(\x_t^i) + r(\x_t^i)\right] - \min_{\x \in \X} \sum_{t=1}^T \left [ f_t(\x) + r(\x) \right]   \leq R(A), \ \forall A \in \A_{\con}.
\end{equation}

\subsection{Proof of Corollary~\ref{cor:com:con}}
Combining Lemma~\ref{lem:smooth} and Theorem~\ref{thm:com:USC:con}, we have
\begin{equation} \label{eqn:com:con:1}
    \begin{split}
        & \sum_{t=1}^T \left [f_t(\x_t) + r(\x_t)\right] - \min_{\x \in \X} \sum_{t=1}^T \left [ f_t(\x) + r(\x) \right]  \\
    \leq &   \min_{A \in \A_{\con}}  R(A)+    \Xi GD   +   \frac{\Psi D}{\sqrt{\ln |\Ept|}} \sqrt{G^2  + 4 H  \sum_{t=1}^T f_t(\x_t)  }.
    \end{split}
\end{equation}
From the theoretical guarantee of COMID for convex and smooth function $f_t(\cdot)$ in Theorem~\ref{thm:COMID:con}, we have
\begin{equation} \label{eqn:com:con:2}
   R(\text{COMID}) \leq \sqrt{8 H D^2}\sqrt{\frac{\delta}{4H}+ \sum_{t=1}^T f_t(\tilde{\x}_t)} + r(\tilde{\x}_1),
\end{equation}
where $\tilde{\x}_t$ denotes the decisions made by COMID. We complete the proof by substituting \eqref{eqn:com:con:2} into \eqref{eqn:com:con:1}:
\begin{equation} \label{eqn:com:con:3}
    \begin{split}
        & \sum_{t=1}^T \left [f_t(\x_t) + r(\x_t)\right] - \min_{\x \in \X} \sum_{t=1}^T \left [ f_t(\x) + r(\x) \right]  \\
    \leq &  \Xi GD + r(\tilde{\x}_1)  +   \frac{\Psi D}{\sqrt{\ln |\Ept|}} \sqrt{G^2  + 4 H  \Lt_{T}^{meta}  } + \sqrt{8 H D^2}\sqrt{\frac{\delta}{4H}+ \Lt_T}
    \end{split}
\end{equation}
where $\Lt_{T}^{meta} = \sum_{t=1}^T f_t(\x_t)$ denotes the cumulative loss  suffered by the meta-algorithm involving $f_t(\cdot)$, and $\Lt_{T}  = \sum_{t=1}^T f_t(\tilde{\x}_t)$ denotes that incurred by COMID.

\section{Conclusion and Future Work}
\label{sec:conclusion}

In this paper, we propose a simple strategy for universal OCO, which can  handle three types of loss functions simultaneously in both the standard and composite settings. The fundamental idea  is to   construct a set of experts by running existing algorithms with different configurations for each type of online functions, and combine them by a meta-algorithm that enjoys a second-order bound with excess losses. The key novelty is to let experts process original functions, and let the meta-algorithm use (partially) linearized losses. In the standard setting, thanks to the second-order bound of the meta-algorithm, our method attains \emph{the best of all worlds} for strongly convex functions and exp-concave functions, up to a double logarithmic factor. For general convex functions, it maintains the minimax optimality and can achieve a small-loss bound. In the composite setting, we employ a different meta-algorithm which is able to achieve a second-order bound that solely depends on the time-varying functions. In this way, our method can  handle multiple types of composite loss functions simultaneously.

There are several directions for future research. First,  our strategy is designed for the purpose of regret minimization, but regret itself may not be suitable for changing environments \citep{IJCAI:2020:Zhang,Online:Review:Casa}. To address this limitation, recent developments in online learning have proposed new performance metrics including adaptive regret \citep{Adaptive:Hazan,Adaptive:ICML:15} and dynamic regret \citep{zinkevich-2003-online,Adaptive:Dynamic:Regret:NIPS}. In the future, we will investigate how to modify our strategy  to support those stronger notions of regret. Second, our strategy needs to fix the value of the time horizon $T$, which is then used to construct $\P_{\str}$  and $\P_{\expc}$. We will study how to design  an \emph{anytime} universal algorithm that does not depend on $T$.

\appendix
\section{Online Extra-gradient Descent (OEGD)} \label{sec:app:OEGD}
In this section, we extend the OEGD algorithm of \citet{Gradual:COLT:12} to strongly convex functions.

\subsection{Algorithm}
There are two sequences of solutions $\{\x_t\}_{t=1}^T$ and $\{\u_t\}_{t=1}^T$, where $\u_t$ is an auxiliary solution used to exploit the smoothness of the loss function.

Based on the property of strong convexity in (\ref{eqn:strong:convex:def}), we set
\[
    \mathcal{R}_t(\x) = \frac{1}{2\eta_t}\|\x\|_2^2
\]
in Algorithm 1 of \citet{Gradual:COLT:12}, where
\begin{equation}\label{eqn:OEGD:etat}
    \eta_t=\frac{2}{\lambda t},
\end{equation}
and obtain the following updating rules:
\begin{equation} \label{eqn:strong:variation}
    \begin{split}
        \u_{t+1}=&\Pi_{\X}\big[\u_t - \eta_t \nabla f_t(\x_t)\big],\\
        \x_{t+1}=&\Pi_{\X}\big[\u_{t+1} - \eta_{t+1} \nabla f_t(\x_t)\big],
    \end{split}
\end{equation}
where $\Pi_{\X}[\cdot]$ denotes the projection onto the nearest point in $\X$, and $\u_1$ and $\x_1$ are set to be any points in $\X$.

\begin{theorem} \label{thm:OEGD:Strong}
    Under Assumptions \ref{ass:1}, \ref{ass:2}, and \ref{ass:3}, if the online functions are $\lambda$-strongly convex, we have
    \[
        \sum_{t=1}^T f_t(\x_t) - \min_{\x \in \X} \sum_{t=1}^T f_t(\x)  \leq \frac{16 G^2}{\lambda} \ln (2 V_T+1) +  \frac{3\lambda D^2}{8}  + \frac{16G^2+4}{\lambda} +   \frac{16 G^2}{\lambda} \ln\left( \frac{256 G^2 H^2}{\lambda^2} +1 \right)
    \]
    where $V_T$ is defined in (\ref{eqn:grad:var}).
\end{theorem}

\subsection{Proof of Theorem~\ref{thm:OEGD:Strong}}
We first introduce a general lemma for optimistic mirror descent \citep[Lemma 3.1]{nemirovski-2005-prox}. 
\begin{lemma}\label{lem:nemirovski}
    Let $\mathcal{Z}$ be a convex compact set in Euclidean space $\mathcal{E}$ with inner product $\langle \cdot, \cdot \rangle$, let $\| \cdot \|$ be a norm on $\mathcal{E}$ and $\|\cdot\|_*$ be its dual norm, and let $\omega(\z): \mathcal{Z} \mapsto \R$ be a $\alpha$-strongly convex function with respect to $\|\cdot\|$, and $\B_\omega(\z,\w)$  be the Bregman distance associated with $\omega$. Let $\mathcal{U}$ be a convex and closed subset of $\mathcal{Z}$, and let $\z_- \in \mathcal{Z}$, let $\bm \xi, \bm \eta \in \mathcal{E}$, and let $\gamma >0$.  Consider the points
    \begin{eqnarray*}
        \w=\argmin_{\y \in \mathcal{U}} \big[\langle \gamma \bm \xi, \y \rangle + \B_\omega(\y, \z_-) \big],\\
        \z_+=\argmin_{\y \in \mathcal{U}} \big[ \langle \gamma \bm \eta , \y \rangle + \B_\omega(\y, \z_-) \big].
    \end{eqnarray*}
    Then for all $\z \in \mathcal{U}$, one has
    \[
         \langle \w - \z,   \gamma \bm \eta \rangle \leq  \B_\omega(\z, \z_-)-\B_\omega(\z,\z_+) + \frac{\gamma^2}{\alpha} \|\bm \eta -\bm \xi \|_*^2 - \frac{\alpha}{2} \big[ \|\w - \z_- \|_2^2 + \| \z_+ - \w \|_2^2 \big]
    \]
    and
    \[
        \| \w - \z_+\| \leq \alpha^{-1} \gamma \|\bm \xi - \bm \eta\|_*.
    \]
\end{lemma}

Applying Lemma~\ref{lem:nemirovski} to the updating rules in (\ref{eqn:strong:variation}), we have
\begin{equation} \label{eqn:mirror}
    \begin{split}
        \langle \nabla f_{t}(\x_{t}),  \x_t - \x \rangle \leq  &  \frac{1}{2 \eta_t} \|\x -\u_{t}\|_2^2 - \frac{1}{2 \eta_t} \|\x -\u_{t+1}\|_2^2 \\
        &+ \eta_t \|\nabla f_{t}(\x_{t}) -\nabla f_{t-1}(\x_{t-1}) \|_2^2 - \frac{1}{2 \eta_t} \big[ \|\x_{t}- \u_{t} \|_2^2 + \| \u_{t+1} - \x_{t} \|_2^2 \big]
    \end{split}
\end{equation}
for any $\x \in \X$ and
\begin{equation}\label{eqn:stability}
    \|\u_{t+1} -\x_{t} \|_2 \leq \eta_t \| \nabla f_{t}(\x_{t}) - \nabla f_{t-1}(\x_{t-1}) \|_2.
\end{equation}
Combining (\ref{eqn:mirror}) with Definition~\ref{def:strong}, we have
\[
    \begin{split}
        f_t(\x_t) -  f_t(\x)  \overset{\eqref{eqn:strong:convex:def},\eqref{eqn:mirror}}{\leq} & \frac{1}{2 \eta_t} \|\x -\u_{t}\|_2^2 - \frac{1}{2 \eta_t} \|\x -\u_{t+1}\|_2^2 - \frac{\lambda}{2}\|\x-\x_t\|_2^2\\
        &+ \eta_t \|\nabla f_{t}(\x_{t}) -\nabla f_{t-1}(\x_{t-1}) \|_2^2 - \frac{1}{2 \eta_t} \big[ \|\x_{t}- \u_{t} \|_2^2 + \| \u_{t+1} - \x_{t} \|_2^2 \big].
    \end{split}
\]
Summing the above inequality over $t=1,\ldots,T$, we have
\begin{equation} \label{eqn:total:bound1}
    \begin{split}
        & \sum_{t=1}^T f_t(\x_t) - \sum_{t=1}^T f_t(\x) \\
        \leq &  \underbrace{\sum_{t=1}^T\frac{\|\x-\u_{t}\|_2^2-\|\x-\u_{t+1}\|_2^2}{2\eta_t}-\frac{\lambda}{2}\|\x-\x_t\|_2^2}_{\ta}\\
        & + \underbrace{\sum_{t=1}^T \eta_t \|\nabla f_{t}(\x_{t}) -\nabla f_{t-1}(\x_{t-1}) \|_2^2 }_{\tb}  -\underbrace{\sum_{t=1}^T\frac{\|\x_t-\u_t\|_2^2+\|\u_{t+1}-\x_t\|_2^2}{2\eta_t}}_{\tc}
    \end{split}	
\end{equation}	
for any $\x \in \X$. In the following, we upper bound the three terms above respectively. For $\ta$, we have
\begin{equation}\label{eqn:bound:a}
    \begin{split}
        \ta = & \frac{1}{2 \eta_1} \|\x -\u_{1}\|_2^2 +  \frac{1}{2} \sum_{t=2}^T \left( \frac{1}{ \eta_t}  - \frac{1}{ \eta_{t-1}} \right) \|\x -\u_{t}\|_2^2 - \frac{1}{2 \eta_T} \|\x -\u_{T+1}\|_2^2\\
        & -\frac{\lambda}{2}\sum_{t=1}^T \|\x - \x_t\|_2^2\\
        \overset{\eqref{eqn:domain}}{\leq}&  \frac{1}{2 \eta_1} D^2 +  \frac{1}{2} \sum_{t=2}^T \left( \frac{1}{ \eta_t}  - \frac{1}{ \eta_{t-1}} \right) \|\x -\u_{t}\|_2^2  -\frac{\lambda}{2}\sum_{t=1}^T \|\x - \x_t\|_2^2\\
        \overset{\eqref{eqn:OEGD:etat}}{=}&  \frac{\lambda D^2}{4}  +  \frac{\lambda}{4} \sum_{t=2}^T  \|\x -\u_{t}\|_2^2  -\frac{\lambda}{2}\sum_{t=1}^T \|\x - \x_t\|_2^2\\ 
        \leq & \frac{\lambda D^2}{4} + \frac{\lambda}{4}\sum_{t=1}^{T-1} \left(\|\x - \u_{t+1}\|_2^2- 2\|\x - \x_t\|_2^2 \right) \leq \frac{\lambda}{4} D^2 + \frac{\lambda}{2}\sum_{t=1}^{T-1} \|\u_{t+1} - \x_t\|_2^2\\ 
        \overset{\eqref{eqn:stability}}{\leq} &  \frac{\lambda D^2}{4}  + \frac{\lambda}{2} \sum_{t=1}^{T-1} \eta_t^2 \| \nabla f_{t}(\x_{t}) - \nabla f_{t-1}(\x_{t-1}) \|_2^2 \\ 
        \leq &  \frac{\lambda D^2}{4}  + \sum_{t=1}^{T-1} \eta_t \| \nabla f_{t}(\x_{t}) - \nabla f_{t-1}(\x_{t-1}) \|_2^2 \leq \frac{\lambda D^2}{4}  + \tb 
    \end{split}
\end{equation}
where in the penultimate line, we use the fact that $\eta_t \leq \eta_1=2/\lambda$.

From (\ref{eqn:bound:a}), we observe that the upper bound of $\ta$ depends on $\tb$. So, we proceed to bound $\tb$. To this end, we define
\[
    \alpha=\left \lceil \sum_{t=1}^T \|\nabla f_t(\x_t)- \nabla f_{t-1}(\x_{t-1})\|_2^2 \right \rceil.
\]
Then, we have
\begin{equation}\label{eqn:bound:b1}
    \begin{split}
        \tb=& \sum_{t=1}^T   \frac{2}{\lambda t} \|\nabla f_t(\x_t)- \nabla f_{t-1}(\x_{t-1})\|_2^2 \\
         = & \sum_{t=1}^\alpha   \frac{2}{\lambda t} \|\nabla f_t(\x_t)- \nabla f_{t-1}(\x_{t-1})\|_2^2 +  \sum_{t=\alpha+1}^T   \frac{2}{\lambda t} \|\nabla f_t(\x_t)- \nabla f_{t-1}(\x_{t-1})\|_2^2\\
        \overset{\eqref{eqn:grad}}{\leq}  & \frac{8G^2}{\lambda} \sum_{t=1}^\alpha \frac{1}{ t}  +  \frac{2}{\lambda (\alpha+1)}\sum_{t=\alpha+1}^T    \|\nabla f_t(\x_t)- \nabla f_{t-1}(\x_{t-1})\|_2^2 \\
        \leq & \frac{8G^2}{\lambda} \left( 1 + \int_{t=1}^\alpha \frac{1}{t} d t \right) + \frac{2}{\lambda} \leq  \frac{8G^2}{\lambda} \left( \ln \alpha +1 \right) + \frac{2}{\lambda} \\
        \leq & \frac{8 G^2}{\lambda} \ln \left(\sum_{t=1}^T \|\nabla f_t(\x_t)- \nabla f_{t-1}(\x_{t-1})\|_2^2 +1 \right) + \frac{8G^2+2}{\lambda}.
    \end{split}
\end{equation}
From  Lemma 12 of \citet{Gradual:COLT:12}, we have
\begin{equation}\label{eqn:variation:trans}
    \sum_{t=1}^T \|\nabla f_t(\x_t)- \nabla f_{t-1}(\x_{t-1})\|_2^2 \leq  2 V_T + 2 H^2 \sum_{t=1}^T \| \x_t -\x_{t-1}\|_2^2.
\end{equation}
Substituting (\ref{eqn:variation:trans}) into (\ref{eqn:bound:b1}), we have 
\begin{equation}\label{eqn:bound:b2}
    \begin{split}
        \tb \overset{\eqref{eqn:bound:b1},\eqref{eqn:variation:trans}}{\leq}  & \frac{8 G^2}{\lambda} \ln \left(2 V_T + 2 H^2 \sum_{t=1}^T \| \x_t -\x_{t-1}\|_2^2+1 \right) + \frac{8G^2+2}{\lambda} \\
        \leq & \frac{8 G^2}{\lambda} \ln (2 V_T+1) +\frac{8 G^2}{\lambda}  \left(2 H^2 \sum_{t=1}^T \| \x_t -\x_{t-1}\|_2^2+1 \right) + \frac{8G^2+2}{\lambda} 
    \end{split}
\end{equation}
where the last step follows from the inequality below
\begin{equation} \label{eqn:sum:log}
    \ln(1+u+v) \leq \ln(1+u)+\ln(1+v), \ \forall u, v \geq 0.
\end{equation}

For  $\tc$, based on the proof of Lemma 21 of \citet{Gradual:COLT:12}, we have
\begin{equation}\label{eqn:bound:c}
	\begin{split}
	        \tc=  \sum_{t=1}^T \frac{\|{\x}_t-\u_t\|_2^2}{2\eta_t}+\sum_{t=2}^{T+1}\frac{\|{\x}_{t-1}-\u_{t}\|_2^2}{2\eta_{t-1}}
		    \geq &  \sum_{t=2}^T \frac{\|{\x}_t-\u_t\|_2^2}{2\eta_{t-1}}+\sum_{t=2}^T \frac{\|{\x}_{t-1}-\u_t\|_2^2}{2\eta_{t-1}}\\
		    \geq & \sum_{t=2}^T \frac{\|{\x}_t-{\x}_{t-1}\|_2^2}{4\eta_{t-1}}\overset{\eqref{eqn:OEGD:etat}}{\geq} \frac{\lambda}{8}\sum_{t=2}^T \|{\x}_t-{\x}_{t-1}\|_2^2.
	\end{split}
\end{equation}
Substituting \eqref{eqn:bound:a}, \eqref{eqn:bound:b2} and \eqref{eqn:bound:c} into \eqref{eqn:total:bound1}, we get
\begin{equation} \label{eqn:total:bound2}
    \begin{split}
        \sum_{t=1}^T f_t(\x_t) - \sum_{t=1}^T f_t(\x) \leq &  \frac{\lambda D^2}{4}  + \frac{16G^2+4}{\lambda} + \frac{16 G^2}{\lambda} \ln (2 V_T+1)\\
        &+\frac{16 G^2}{\lambda}  \left(2 H^2 \sum_{t=1}^T \| \x_t -\x_{t-1}\|_2^2+1 \right) - \frac{\lambda}{8}\sum_{t=2}^T \|{\x}_t-{\x}_{t-1}\|_2^2 \\
        \overset{\eqref{eqn:domain}}{\leq}& \frac{3\lambda D^2}{8}  + \frac{16G^2+4}{\lambda} + \frac{16 G^2}{\lambda} \ln (2 V_T+1)\\
        &+\frac{16 G^2}{\lambda}  \left(2 H^2 \sum_{t=1}^T \| \x_t -\x_{t-1}\|_2^2+1 \right) - \frac{\lambda}{8}\sum_{t=1}^T \|{\x}_t-{\x}_{t-1}\|_2^2.
    \end{split}	
\end{equation}

To simplify the above inequality, we make use of the following inequality \citep[Lemma 7]{ICML:2023:Chen}.
\begin{lemma} \label{lem:ineq} 
    Let $A\geq 0$, $a\geq0$, $b \geq 0$ and $c>0$, we have
    \[
        a \ln(b A+1) - c A \leq a \ln\left( \frac{ab}{c}+1\right).
    \]
\end{lemma}
From Lemma~\ref{lem:ineq}, we have
\begin{equation} \label{eqn:log:diff}
    \frac{16 G^2}{\lambda}  \left(2 H^2 \sum_{t=1}^T \| \x_t -\x_{t-1}\|_2^2+1 \right) - \frac{\lambda}{8}\sum_{t=1}^T \|{\x}_t-{\x}_{t-1}\|_2^2 \leq \frac{16 G^2}{\lambda}\ln \left( \frac{256 G^2 H^2}{\lambda^2} +1 \right).
\end{equation}
Combining (\ref{eqn:total:bound2}) with (\ref{eqn:log:diff}), we complete the proof.

\section{Composite Objective Mirror Descent (COMID)}
\label{sec:app:COMID}
In this section, we first revisit the COMID algorithm of \cite{COLT:2010:Composite}, and then establish (pseudo) small-loss bounds for general convex $f_t(\cdot)$ and strongly convex $f_t(\cdot)$.

\subsection{Algorithm}
The updating rule of COMID \citep{COLT:2010:Composite} is shown below 
\begin{equation} \label{eqn:comid}
    \x_{t+1} = \argmin_{\x \in \X} \left\{   \left<\nabla f_t(\x_t), \x \right> +  \B_{\mathcal{R}_t}(\x, \x_t) +  r(\x) \right\},
\end{equation}
where  $\mathcal{R}_t$ is a strongly convex function and $\B_{\mathcal{R}_t}$ is the Bregman distance associated with $\mathcal{R}_t$. In our setting, we set 
\[
    \mathcal{R}_t(\x) = \frac{1}{2 \eta_t}\|\x\|_2^2.
\] 
Then, the original updating rule \eqref{eqn:comid} of COMID becomes:
\begin{equation} 
\begin{split}
\x_{t+1} = \argmin_{\x \in \X} \left\{ \left<\nabla f_t(\x_t), \x \right> + \frac{1}{2\eta_t }\|\x - \x_t\|_2^2 +   r(\x) \right\}.
\end{split}
\end{equation}

For the general convex function $f_t(\cdot)$, we set the learning rate as below 
\begin{equation}\label{eqn:COMID:etat}
\eta_t= \frac{\alpha}{\sqrt{\delta + \sum_{s=1}^{t}\|\nabla f_s(\x_s)\|_2^2}},
\end{equation}
where the parameter $\delta > 0$ is introduced to avoid being divided by $0$, and $\alpha > 0 $ is used to fine-tune the upper bound. Under the smoothness condition in Assumption~\ref{ass:3}, COMID ensures the following theoretical guarantee.
\begin{theorem} \label{thm:COMID:con}
    Under Assumptions \ref{ass:1}, \ref{ass:2}, \ref{ass:3}, \ref{ass:4} and \ref{ass:5}, if the time-varying function $f_t(\cdot)$ is general convex, we have
    \begin{equation}
         \sum_{t=1}^T \big[ f_t(\x_t) + r(\x_t) \big] - \sum_{t=1}^T \big[f_t(\x) + r(\x) \big] \leq \sqrt{8 H D^2}\sqrt{\frac{\delta}{4H}+ \Lt_T} + C =  O \left(\sqrt{\Lt_T} \right), \nonumber
    \end{equation}
    where $\Lt_T$ is defined in \eqref{eqn:L:til}.
\end{theorem}

For the strongly convex  function $f_t(\cdot)$, we modify the learning rate as:
\begin{equation}\label{eqn:COMID:etat:str}
    \eta_t= \frac{2}{ \lambda t}, 
\end{equation}
and obtain the following regret bound.
\begin{theorem} \label{thm:COMID:str}
    Under Assumptions \ref{ass:1}, \ref{ass:2}, \ref{ass:3}, \ref{ass:4} and \ref{ass:5}, if the time-varying function $f_t(\cdot)$ is $\lambda$-strongly convex, we have
    \begin{equation}
        \begin{split}
             \sum_{t=1}^T \big[ f_t(\x_t) + r(\x_t) \big] - \min_{\x \in \X} \sum_{t=1}^T \big[f_t(\x) + r(\x) \big]  \leq &   \frac{\lambda D^2}{4} + C  + \frac{1 + G^2}{\lambda}  +  \frac{G^2}{2\lambda}  \ln \left( 4H \Lt_T + 1\right) \\
             = & O \left( \frac{1}{\lambda}\log \Lt_T \right), \nonumber
        \end{split}
    \end{equation}
    where $\Lt_T$ is defined in \eqref{eqn:L:til}.
\end{theorem}

\subsection{Proof of Theorem~\ref{thm:COMID:con}}
Let $\x^* = \argmin_{\x \in \X} \sum_{t=1}^T [f_t(\x) + r(\x)]$. From the analysis of \citet[Theorem 2]{COLT:2010:Composite}, we have 
\begin{equation}
    \begin{split}
        & \sum_{t=1}^T \big[ f_t(\x_t) + r(\x_t) \big] - \min_{\x \in \X} \sum_{t=1}^T \big[f_t(\x) + r(\x) \big] \\
         \leq & \sum_{t=1}^T \left\{\frac{1}{2\eta_t} \left[ \|\x_t - \x^*\|_2^2 - \|\x_{t+1} - \x^*\|_2^2 \right] + \frac{\eta_t}{2}\|\nabla f_t(\x_t)\|_2^2\right\} + r(\x_1). \nonumber
    \end{split}
\end{equation}
Under Assumptions~\ref{ass:2}~and~\ref{ass:5}, the above result can be simplified as
\begin{equation}    \label{eqn:COMID:1}
    \begin{split}
         & \sum_{t=1}^T \big[ f_t(\x_t) + r(\x_t) \big] - \min_{\x \in \X} \sum_{t=1}^T \big[f_t(\x) + r(\x) \big] \\
         \leq & \frac{1}{2\eta_1}  \|\x_{1} - \x^*\|_2^2 + \sum_{t=2}^T\left(\frac{1}{2\eta_t} - \frac{1}{2\eta_{t-1}}\right)\|\x_{t} - \x^*\|_2^2 + \sum_{t=1}^T \frac{\eta_t}{2} \|\nabla f_t(\x_t)\|_2^2 + C \\
         \leq & \frac{D^2}{2\eta_1}   + D^2\sum_{t=2}^T\left(\frac{1}{2\eta_t} - \frac{1}{2\eta_{t-1}}\right) + \sum_{t=1}^T \frac{\eta_t}{2} \|\nabla f_t(\x_t)\|_2^2 + C \\
         =& \frac{D^2}{2\eta_T}  + \sum_{t=1}^T \frac{\eta_t}{2} \|\nabla f_t(\x_t)\|_2^2 + C.
    \end{split}
\end{equation}
To upper bound $\sum_{t=1}^T \frac{\eta_t}{2} \|\nabla f_t(\x_t)\|_2^2$  in \eqref{eqn:COMID:1}, we make use of the following lemma \citep[Lemma 3.5]{AUER200248}.

\begin{lemma} \label{lem:sum} 
    Let $l_1,\ldots$, $l_T$ and $\delta$ be non-negative real numbers. Then
    \[
        \sum_{t=1}^T \frac{l_t}{\sqrt{\delta+\sum_{s=1}^t l_s}} \leq 2 \left( \sqrt{\delta + \sum_{t=1}^T l_t} -\sqrt{\delta}\right)
    \]
    where $0/\sqrt{0}=0$.
\end{lemma}
Applying Lemma~\ref{lem:sum}, we have
\begin{equation}\label{eqn:COMID:2}
    \begin{split}
        \sum_{t=1}^T \frac{\eta_t}{2}\|\nabla f_t(\x_{t})\|_2^2  \overset{\eqref{eqn:COMID:etat}}{=}   \frac{\alpha}{2} \sum_{t=1}^T   \frac{ \|\nabla f_t(\x_{t})\|_2^2}{\sqrt{\delta+ \sum_{s=1}^t \|\nabla f_s(\x_{s})\|_2^2} }
        \leq \alpha  \sqrt{\delta+ \sum_{t=1}^T \|\nabla f_t(\x_{t})\|_2^2}.
    \end{split}
\end{equation}
Substituting \eqref{eqn:COMID:2} into \eqref{eqn:COMID:1}, we have
\begin{equation}    \label{eqn:COMID:3}
    \begin{split}
         \sum_{t=1}^T \big[ f_t(\x_t) + r(\x_t) \big] - \min_{\x \in \X} \sum_{t=1}^T \big[f_t(\x) + r(\x) \big]  \leq &  \left( \frac{D^2}{2 \alpha} +\alpha \right) \sqrt{\delta+ \sum_{t=1}^T \|\nabla f_t(\x_{t})\|_2^2} + C \\
         = & \sqrt{2D^2} \sqrt{\delta+ \sum_{t=1}^T \|\nabla f_t(\x_{t})\|_2^2} + C,
    \end{split}
\end{equation}
where  $\alpha= D/\sqrt{2}$.

We finish the proof by making use of Lemma~\ref{lem:smooth}:
\begin{equation}
    \begin{split}
        &\sum_{t=1}^T \big[ f_t(\x_t) + r(\x_t) \big] - \min_{\x \in \X} \sum_{t=1}^T \big[f_t(\x) + r(\x) \big] \\
        \overset{\eqref{eqn:smoothness:property},\eqref{eqn:COMID:3}}{\leq}&  \sqrt{2D^2} \sqrt{ \delta+ 4 H \sum_{t=1}^T  f_t(\x_{t}) } + C=\sqrt{8 H D^2}\sqrt{\frac{\delta}{4H}+ \sum_{t=1}^T f_t(\x_t)} + C.
    \end{split}
\end{equation}

\subsection{Proof of Theorem~\ref{thm:COMID:str}}
From the analysis of \citet[Theorem 7]{COLT:2010:Composite}, we have 
\begin{equation}
    \begin{split}
        & \sum_{t=1}^T \big[ f_t(\x_t) + r(\x_t) \big] - \min_{\x \in \X} \sum_{t=1}^T \big[f_t(\x) + r(\x) \big] \\
        \leq & \sum_{t=1}^T \left\{\frac{1}{2\eta_t} \left[ \|\x_t - \x^*\|_2^2 - \|\x_{t+1} - \x^*\|_2^2 \right] + \frac{\eta_t}{2}\|\nabla f_t(\x_t)\|_2^2 - \frac{\lambda}{2}\|\x_t - \x^*\|_2^2\right\} + r(\x_1), \nonumber
    \end{split}
\end{equation}
where $\x^* = \argmin_{\x \in \X} \sum_{t=1}^T [f_t(\x) + r(\x)]$. Then, we make use of Assumptions~\ref{ass:2}~and~\ref{ass:5} to simplify the above inequality
\begin{equation}   \label{eqn:COMID:str:1}
    \begin{split}
         & \sum_{t=1}^T \big[ f_t(\x_t) + r(\x_t) \big] - \min_{\x \in \X} \sum_{t=1}^T \big[f_t(\x) + r(\x) \big] \\
         \leq & \frac{D^2}{2\eta_1}  + \sum_{t=2}^T\left(\frac{1}{2\eta_t} - \frac{1}{2\eta_{t-1}} - \frac{\lambda}{2}\right)\|\x_{t} - \x^*\|_2^2 + \sum_{t=1}^T \frac{\eta_t}{2} \|\nabla f_t(\x_t)\|_2^2 + C.
    \end{split}
\end{equation}
According to \eqref{eqn:COMID:etat:str}, we have 
\begin{equation} \label{eqn:COMID:str:2}
    \frac{1}{\eta_t} - \frac{1}{\eta_{t-1}} - \lambda = \frac{\lambda }{2} - \lambda  \leq  0.
\end{equation}
Substituting \eqref{eqn:COMID:str:2} into \eqref{eqn:COMID:str:1}, we get
\begin{equation}    \label{eqn:COMID:str:3}
    \begin{split}
         \sum_{t=1}^T \big[ f_t(\x_t) + r(\x_t) \big] - \min_{\x \in \X} \sum_{t=1}^T \big[f_t(\x) + r(\x) \big]  \leq  &  \frac{D^2}{2\eta_1}    + C + \sum_{t=1}^T \frac{\eta_t}{2} \|\nabla f_t(\x_t)\|_2^2 \\
         =& \frac{\lambda D^2}{4} + C + \sum_{t=1}^T \frac{1}{\lambda t} \|\nabla f_t(\x_t)\|_2^2 . 
    \end{split}
\end{equation}
Then, we proceed to bound $\sum_{t=1}^T \frac{1}{\lambda t} \|\nabla f_t(\x_t)\|_2^2$. To this end, we define 
\[
    \alpha=\left \lceil \sum_{t=1}^T \|\nabla f_t(\x_t)\|_2^2 \right \rceil.
\]
Next, we have 
\begin{equation}    \label{eqn:COMID:str:4}
    \begin{split}
        \sum_{t=1}^T \frac{1}{\lambda t} \|\nabla f_t(\x_t)\|_2^2 = &  \sum_{t=1}^\alpha \frac{1}{\lambda t} \|\nabla f_t(\x_t)\|_2^2 + \sum_{t=\alpha + 1}^T \frac{1}{\lambda t} \|\nabla f_t(\x_t)\|_2^2 \\
        \overset{\eqref{eqn:grad}}{\leq} & \frac{G^2}{\lambda} \sum_{t=1}^\alpha  \frac{1}{t} + \frac{1}{\lambda (\alpha + 1)} \sum_{t=\alpha + 1}^T  \|\nabla f_t(\x_t)\|_2^2 \\
        \leq & \frac{G^2}{\lambda} \left( 1 + \int_{t=1}^\alpha \frac{1}{t} d t \right) + \frac{1}{\lambda} \leq \frac{G^2}{\lambda} \left(\ln \alpha + 1\right) + \frac{1}{\lambda} \\
        \leq &  \frac{G^2}{\lambda}  \ln \left( \sum_{t=1}^T \|\nabla f_t(\x_t)\|_2^2 + 1 \right)  + \frac{1 + G^2}{\lambda}.
    \end{split}
\end{equation}

Substituting \eqref{eqn:COMID:str:4} into \eqref{eqn:COMID:str:3}, we obtain
\begin{equation}     \label{eqn:COMID:str:5}
    \begin{split}
         & \sum_{t=1}^T \big[ f_t(\x_t) + r(\x_t) \big] - \min_{\x \in \X} \sum_{t=1}^T \big[f_t(\x) + r(\x) \big] \\
         \leq  &  \frac{\lambda D^2}{4} + C  + \frac{1 + G^2}{\lambda}  +  \frac{G^2}{2\lambda}  \ln \left( \sum_{t=1}^T \|\nabla f_t(\x_t)\|_2^2 + 1\right).
    \end{split}
\end{equation}  
Combining \eqref{eqn:COMID:str:5}  and Lemma~\ref{lem:smooth} completes the proof.

\section{Optimistic Composite Mirror Descent (OCMD)} 
\label{sec:app:optcmd}
In this section, we revisit the OCMD algorithm of \cite{TAC:2023:Scroccaro}, and then extend OCMD to the exp-concave case, i.e.,~$f_t(\cdot)$ is exp-concave.

\subsection{Algorithm}
During the online process, OCMD maintains two sequences of solutions $\{\x_t\}_{t=1}^T$ and $\{\u_t\}_{t=1}^T$, where $\u_t$ is an auxiliary solution used to exploit the smoothness of the loss function. The updating rules are shown below:
\begin{equation}  
    \begin{split}
        \u_{t+1} =& \argmin_{\x \in \X}\left\{ \left \langle \nabla f_t(\x_t), \x \right \rangle + r(\x) + \B_{\mathcal{R}_{t}}(\x, \u_t)  \right\},  \\    
        \x_{t+1} = & \argmin_{\x \in \X}\left\{ \left \langle M_{t+1}, \x \right \rangle + r(\x) + \B_{\mathcal{R}_{t+1}}(\x, \u_{t+1})  \right\},
    \end{split}
\end{equation}
where $M_{t+1}$ denotes the optimistic estimation for  the gradient of $f_{t+1}(\cdot)$ and is typically set as $M_{t+1} = \nabla f_t(\x_t)$ to exploit the smoothness of $f_t(\cdot)$.

To deal with the $\alpha$-exp-concave $f_t(\cdot)$, we follow \citet{Gradual:COLT:12} and  set 
\begin{equation}    \label{eq:R:exp}
    \mathcal{R}_t(\x)  = \frac{1}{2}\|\x\|_{H_t}^2
\end{equation}
where $H_t = I + (\beta  G^2/2) I +  (\beta   /2) \sum_{s=1}^{t-1} h_s$, $I$ denotes the $d$-dimension identity matrix, $h_t=\nabla f_t (\x_t) \nabla f_t (\x_t)^\top$  and $\beta = (1/2)\min \{1/(4GD), \alpha\}$.

With the above configurations, we can obtain the following gradient-variation bound of OCMD for exp-concave $f_t(\cdot)$.
\begin{theorem}    \label{thm:OCMD:exp:VT}
     Under Assumptions \ref{ass:1}, \ref{ass:2}, \ref{ass:3}, \ref{ass:4} and \ref{ass:5}, if the time-varying function $f_t(\cdot)$ is $\alpha$-exp-concave, we have
    \begin{equation}
         \sum_{t=1}^T \big[ f_t(\x_t) + r(\x_t) \big] - \sum_{t=1}^T \big[f_t(\x) + r(\x) \big] \leq  O \left(\frac{d}{\alpha} \log V_T \right), \nonumber
    \end{equation}
    where  $V_T$ is defined in \eqref{eqn:grad:var}.
\end{theorem}

Furthermore, we establish the following (pseudo) small-loss bound.
\begin{theorem}    \label{thm:OCMD:exp:LT}
     Under Assumptions \ref{ass:1}, \ref{ass:2}, \ref{ass:3}, \ref{ass:4} and \ref{ass:5}, if the time-varying function $f_t(\cdot)$ is $\alpha$-exp-concave, we have
    \begin{equation}
         \sum_{t=1}^T \big[ f_t(\x_t) + r(\x_t) \big] - \sum_{t=1}^T \big[f_t(\x) + r(\x) \big] \leq  O \left(\frac{d}{\alpha} \log \Lt_T \right), \nonumber
    \end{equation}
    where  $\Lt_T$ is defined in \eqref{eqn:L:til}.
\end{theorem}

\subsection{Proof of Theorem~\ref{thm:OCMD:exp:VT}}

We present the following lemma regarding  OCMD  \citep{TAC:2023:Scroccaro}, which can be treated as an extension of Lemma~\ref{lem:nemirovski}  to composite optimization.

\begin{lemma}    \label{lem:ocmd}
    Under the conditions of Lemma~\ref{lem:nemirovski}, and assume $r(\cdot)$ is convex.  Consider the points
    \begin{eqnarray*}
        \w=\argmin_{\y \in \mathcal{U}} \big[\langle \gamma \bm \xi, \y \rangle + r(\y) + \B_\omega(\y, \z_-) \big],\\
        \z_+=\argmin_{\y \in \mathcal{U}} \big[ \langle \gamma \bm \eta , \y \rangle + r(\y) +\B_\omega(\y, \z_-) \big].
    \end{eqnarray*}
    Then for all $\z \in \mathcal{U}$, one has
    \begin{align*}
        {}&\left\langle \w - \z, \gamma {\bm \eta} \right\rangle + r(\w)-r(\z) \\
        \leq{}& \frac{\gamma^2}{\alpha}\| \bm \xi - \bm \eta\|_* + \B_{\omega}(\z,\z_{-}) - \B_{\omega}(\z,\z_+)- \frac{\alpha}{2} \big[ \|\w - \z_+ \|_2^2 + \| \z_- - \w \|_2^2 \big].
    \end{align*}
    and
    \[
        \| \w - \z_+\| \leq \alpha^{-1} \gamma \|\bm \xi - \bm \eta\|_*.
    \]
\end{lemma}

\noindent
Let $\x^* = \argmin_{\x \in \X} \sum_{t=1}^T [f_t(\x) + r(\x)]$. According to Lemma~\ref{lem:exp},  we have 
\begin{equation}
    \begin{split}
        & \sum_{t=1}^T \left[f_t(\x_t) + r(\x_t) \right] -   \sum_{t=1}^T \left[f_t(\x^*) + r(\x^*) \right]  \\
        \leq &   \sum_{t=1}^T \left[ \left< \nabla f_t(\x_t), \x_t - \x^* \right> + r(\x_t) - r(\x^*) - \frac{\beta}{2} \left<\nabla f_t(\x_t), \x^* - \x_t \right>^2 \right]. \nonumber
    \end{split}
\end{equation}
Then, applying Lemma~\ref{lem:ocmd} with $ {\bm \xi}= \nabla f_{t-1}(\x_{t-1}) ,  {\bm \eta} = \nabla f_t(\x_t)$, $\alpha = \gamma = 1$, and $\omega(\cdot) = \mathcal{R}_t(\cdot)$ shown in \eqref{eq:R:exp},  we obtain 
\begin{equation}    \label{eqn:ocmd:x-u}
    \|\x_t - \u_{t+1}\|_{H_t} \leq \|\nabla f_{t-1}(\x_{t-1}) - \nabla f_{t}(\x_{t})\|_{H_t^{-1}}
\end{equation}
and 
\begin{equation}
    \begin{split}
        &\sum_{t=1}^T \left[ \langle \nabla f_t (\x_t), \x_{t} - \x^*\rangle + r(\x_t) - r(\x^*)   - \frac{\beta}{2}   \|\x^* - \x_t\|_{h_t}^2 \right] 
        \leq \underbrace{  \sum_{t=1}^T    \| \nabla f_t(\x_t) - \nabla f_{t-1}(\x_{t-1})  \|_{H_t^{-1}}^2}_{\ta}\\
        & +  \underbrace{  \frac{1}{2}  \sum_{t=1}^T \left[  \|\x^* - \u_t\|_{H_t}^2 - \|\x^* - \u_{t+1}\|_{H_t}^2 - \beta \|\x^* - \x_t\|_{h_t}^2 \right]}_{\tb} - \underbrace{ \frac{1}{2}\sum_{t=1}^T   \left[  \|\u_{t+1} - \x_t\|_{H_t}^2 + \|\x_{t} - \u_t\|_{H_t}^2\right]  }_{\tc}. \nonumber
    \end{split}
\end{equation}
Next, we analyze the above three terms separately.  
For $\ta$, we will utilize the following inequality:
\begin{equation}    \label{eq:exp:Ht}
    \begin{split}
        H_{t} \succeq &  I+\frac{\beta}{4} \sum_{\tau=1}^{t}\left(\nabla f_{\tau}\left(\x_{\tau}\right) \nabla f_{\tau}\left(\x_{\tau}\right)^{\top}+\nabla f_{\tau-1}\left(\x_{\tau-1}\right) \nabla f_{\tau-1}\left(\x_{\tau-1}\right)^{\top}\right) \\
         \succeq & I+\frac{\beta}{8} \sum_{\tau=1}^{t}\left(\nabla f_{\tau}\left(\x_{\tau}\right)-\nabla f_{\tau-1}\left(\x_{\tau-1}\right)\right)\left(\nabla f_{\tau}\left(\x_{\tau}\right)-\nabla f_{\tau-1}\left(\x_{\tau-1}\right)\right)^{\top} = P_t,
    \end{split}
\end{equation}
where the first step is due to  $G^2 I \succeq \nabla f_{t}\left(\x_{t}\right) \nabla f_{t}\left(\x_{t}\right)^{\top}$, and the second step is due to 
\begin{equation}
    \begin{split}
        \nabla f_{\tau}\left(\x_{\tau}\right) \nabla f_{\tau}\left(\x_{\tau}\right)^{\top} & +\nabla f_{\tau-1}\left(\x_{\tau-1}\right) \nabla f_{\tau-1}\left(\x_{\tau-1}\right)^{\top} \\ & \succeq \frac{1}{2}\left(\nabla f_{\tau}\left(\x_{\tau}\right)-\nabla f_{\tau-1}\left(\x_{\tau-1}\right)\right)\left(\nabla f_{\tau}\left(\x_{\tau}\right)-\nabla f_{\tau-1}\left(\x_{\tau-1}\right)\right)^{\top}. \nonumber
    \end{split}
\end{equation}
According to \eqref{eq:exp:Ht}, we have
\begin{equation}       \label{eq:exp:ta:1}
    \begin{split}
        \ta \leq & \sum_{t=1}^T    \| \nabla f_t(\x_t) - \nabla f_{t-1}(\x_{t-1})  \|_{P_t^{-1}}^2  
        = \frac{8}{\beta} \sum_{t=1}^{T}\left\|\sqrt{\frac{\beta}{8}}\left(\nabla f_{t}\left(\x_{t}\right)-\nabla f_{t-1}\left(\x_{t-1}\right)\right)\right\|_{P_{t}^{-1}}^{2}.  
    \end{split}
\end{equation}
Then, we introduce the following lemma \citep[Lemma 10]{ICML:2023:Chen}.
\begin{lemma}      \label{lem:chen}
    Let $u_t \in \mathbb{R}^d$  $( t = 1, \ldots, T $, be a sequence of vectors. Define $ S_t = \sum_{i=1}^t u_i u_i^\top + \epsilon I,$ where $ \epsilon > 0 $. Then $ \sum_{t=1}^T u_t^\top S_t^{-1} u_t \leq d \ln \left( 1 + \frac{1}{d\epsilon} \sum_{t=1}^T \|u_t\|_2^2 \right)$. 
\end{lemma}
Applying Lemma~\ref{lem:chen} with $\u_t = \sqrt{\frac{\beta}{8}}(\nabla f_{t}\left(\x_{t}\right)-\nabla f_{t-1} (\x_{t-1}))$ and $\epsilon = 1$, we have 
\begin{equation}    \label{eq:exp:ta:2}
    \begin{split}
         \sum_{t=1}^{T}\left\|\sqrt{\frac{\beta}{8}}\left(\nabla f_{t}\left(\x_{t}\right)-\nabla f_{t-1}\left(\x_{t-1}\right)\right)\right\|_{P_{t}^{-1}}^{2}   
         \leq   d \ln \left(\frac{\beta}{8 d} \bar{V}_{T}+1\right),  
    \end{split}
\end{equation}
where we define
\begin{equation}    \label{eqn:bar:VT}
    \bar{V}_{T} = \sum_{t=1}^T \|\nabla f_{t}\left(\x_{t}\right)-\nabla f_{t-1} (\x_{t-1})\|_2^2.
\end{equation}
Combining \eqref{eq:exp:ta:1} and \eqref{eq:exp:ta:2}, we arrive at
\begin{equation}    \label{eq:exp:ta}
    \begin{split}
        \ta \leq &     \frac{8 d}{\beta} \ln \left(\frac{\beta}{8 d} \bar{V}_{T}+1\right).
    \end{split}
\end{equation}

For $\tb$, we exploit the fact that $H_{t+1} - H_t = \frac{\beta}{2}h_t$ and obtain
\begin{align}        \label{eq:exp:tb-1}
        \tb =& \frac{1}{2} \left[ \|\x^* - \u_1\|_{H_1}^2 - \|\x^* - \u_{T+1}\|_{H_{T+1}}^2\right] \nonumber \\ 
         &   + \frac{1}{2}\sum_{t=1}^T \left\{ \|\x^* - \u_{t+1}\|_{H_{t+1}}^2 - \|\x^* - \u_{t+1}\|_{H_t}^2 -  \beta  \|\x^* - \x_t\|_{h_t}^2 \right \} \nonumber \\
        =& \frac{1}{2} \left[  \|\x^* - \u_1\|_{H_1}^2 - \|\x^* - \u_{T+1}\|_{H_{T+1}}^2 \right] + \frac{\beta}{4} \sum_{t=1}^T \left\{  \|\x^* - \u_{t+1}\|_{h_t}^2 - 2\|\x^* - \x_t\|_{h_t}^2 \right \} \nonumber \\
         \overset{\eqref{eqn:grad},\eqref{eqn:domain}}{\leq}& \left(\frac{1}{2} + \frac{\beta}{4} G^2\right) D^2 + \frac{\beta}{4} \sum_{t=1}^T \left\{  \|\x^* - \u_{t+1}\|_{h_t}^2 - 2 \|\x^* - \x_t\|_{h_t}^2 \right \}.
\end{align}
According to the fact that  $H_t \succeq \frac{\beta}{2} G^2 I \succeq \frac{\beta}{2} h_t$ and \eqref{eqn:ocmd:x-u}, we have 
\begin{equation}    \label{eq:exp:tb-2}
    \begin{split}
         & \|\x^* - \u_{t+1}\|_{h_t}^2 - 2  \|\x^* - \x_t\|_{h_t}^2   \leq 2    \|\x_t - \u_{t+1}\|_{h_t}^2 \\
         \leq &  \frac{4}{\beta } \|\x_t - \u_{t+1}\|_{H_t}^2 
         \overset{\eqref{eqn:ocmd:x-u}}{\leq} \frac{4}{\beta } \|\nabla f_t(\x_t) - \nabla f_{t-1}(\x_{t-1})\|_{H_t^{-1}}.
    \end{split}
\end{equation}
Substituting \eqref{eq:exp:tb-2} into \eqref{eq:exp:tb-1}, we have 
\begin{align}       \label{eq:exp:tb}
        \tb \leq \left(\frac{1}{2} + \frac{\beta}{4} G^2\right) D^2 +  \ta.
\end{align}

For $\tc$, we have 
\begin{equation}     \label{eq:exp:tc}
    \begin{split}
        \tc &= \frac{1}{2} \sum_{t=2}^{T+1}  \|\u_{t} - \x_{t-1}\|_{H_{t-1}}^2  + \frac{1}{2} \sum_{t=1}^T  \|\x_{t} - \u_t\|_{H_t}^2  \\
        &\geq \frac{1}{2} \sum_{t=2}^T \left\{  \|\u_{t} - \x_{t-1}\|_2^2 + \|\x_{t} - \u_t\|_2^2 \right \} \geq \frac{1}{4}  \sum_{t=2}^T \|  \x_t  -  \x_{t-1}  \|_2^2,
    \end{split}
\end{equation}
where the first inequality is due to $H_{t} \succeq H_{t-1} \succeq  I$, $\|\u_{T+1} - \x_{T}\|_{H_{T+1}}^2 \geq 0$ and  $\|\x_{1} - \u_{1}\|_{H_{1}}^2 \geq 0$.

Combining \eqref{eq:exp:ta}, \eqref{eq:exp:tb} and \eqref{eq:exp:tc}, we have 
\begin{equation}    \label{eq:exp:optcmd:1}
    \begin{split}
        & \sum_{t=1}^T \left[f_t(\x_t) + r(\x_t) \right] -   \sum_{t=1}^T \left[f_t(\x^*) + r(\x^*) \right]  \\
        \leq &  \left(\frac{1}{2} + \frac{\beta}{4} G^2\right) D^2 +  \frac{16 d}{\beta} \ln \left(\frac{\beta}{8 d} \bar{V}_{T}+1\right) - \frac{1}{4}  \sum_{t=2}^T \|  \x_t  -  \x_{t-1}  \|_2^2.
    \end{split}
\end{equation}
Notice that $\bar{V}_T$ in \eqref{eqn:bar:VT} can be bounded in the following way: 
\begin{equation}    \label{eq:exp:optcmd:2}
    \begin{split}
         \bar{V}_T & \leq G^{2}+2 \sum_{t=2}^{T}\left\|\nabla f_{t}\left(\mathbf{x}_{t}\right)-\nabla f_{t-1}\left(\mathbf{x}_{t}\right)\right\|_{2}^{2}+2 \sum_{t=2}^{T}\left\|\nabla f_{t-1}\left(\mathbf{x}_{t}\right)-\nabla f_{t-1}\left(\mathbf{x}_{t-1}\right)\right\|_{2}^{2} \\ 
         & \leq G^{2}+2V_T+2 H^{2} \sum_{t=2}^{T}\left\|\mathbf{x}_{t}-\mathbf{x}_{t-1}\right\|_{2}^{2}.
    \end{split}
\end{equation}
Substituting \eqref{eq:exp:optcmd:2} into \eqref{eq:exp:optcmd:1}, we have
\begin{align}
        & \sum_{t=1}^T \left[f_t(\x_t) + r(\x_t) \right] - \sum_{t=1}^T \left[f_t(\x^*) + r(\x^*) \right]  
        \leq \left(\frac{1}{2} + \frac{\beta}{4} G^2\right) D^2  - \frac{1}{4}  \sum_{t=2}^T \|  \x_t  -  \x_{t-1}  \|_2^2 \nonumber \\ 
        & +  \frac{16 d}{\beta} \ln \left(\frac{\beta}{8 d} G^2 + \frac{\beta}{4 d} V_T + \frac{\beta H^2}{4 d} \sum_{t=2}^{T}\left\|\mathbf{x}_{t}-\mathbf{x}_{t-1}\right\|_{2}^{2} +1\right) \nonumber \\ 
        \overset{\eqref{eqn:sum:log}}{\leq} & \left(\frac{1}{2} + \frac{\beta}{4} G^2\right) D^2  + \frac{16 d}{\beta} \ln \left(\frac{\beta}{8 d} G^2 + \frac{\beta}{4 d} V_T   +1\right) \nonumber \\ 
        & + \frac{16 d}{\beta} \ln \left( \frac{\beta H^2}{4 d} \sum_{t=2}^{T}\left\|\mathbf{x}_{t}-\mathbf{x}_{t-1}\right\|_{2}^{2} +1\right) - \frac{1}{4}  \sum_{t=2}^T \|  \x_t  -  \x_{t-1}  \|_2^2. \label{eq:exp:optcmd:3}
\end{align}

To simplify \eqref{eq:exp:optcmd:3}, we make use of Lemma~\ref{lem:ineq} and obtain
\begin{equation} \label{eq:exp:optcmd:4}
\frac{16 d}{\beta} \ln  \left(\frac{\beta H^2}{4d} \sum_{t=2}^T \| \x_t -\x_{t-1}\|_2^2+1 \right) - \frac{1}{4}\sum_{t=2}^T \|{\x}_t-{\x}_{t-1}\|_2^2 \leq \frac{16 d}{\beta} \ln \left( \frac{\beta H^2}{d} +1 \right).
\end{equation}
Combining \eqref{eq:exp:optcmd:3} and \eqref{eq:exp:optcmd:4}, we complete the proof.

\subsection{Proof of Theorem~\ref{thm:OCMD:exp:LT}}
Notice that for $\bar{V}_T$ in \eqref{eqn:bar:VT}, we have 
\begin{equation}    \label{eq:exp:optcmd:5}
    \begin{split}
        \bar{V}_{T} & \leq\left\|\nabla f_{1}\left(\mathbf{x}_{1}\right)\right\|_{2}^{2}+2 \sum_{t=2}^{T}\left\|\nabla f_{t}\left(\mathbf{x}_{t}\right)\right\|_{2}^{2}+2 \sum_{t=2}^{T}\left\|\nabla f_{t-1}\left(\mathbf{x}_{t-1}\right)\right\|_{2}^{2} \\
         & \leq 8 H \sum_{t=1}^{T} f_{t}\left(\mathbf{x}_{t}\right)+8 H \sum_{t=2}^{T} f_{t-1}\left(\mathbf{x}_{t-1}\right) \leq 16 H \sum_{t=1}^{T} f_{t}\left(\mathbf{x}_{t}\right),
    \end{split}
\end{equation}
where the second step is due to Lemma~\ref{lem:smooth}.
Substituting \eqref{eq:exp:optcmd:5} into \eqref{eq:exp:optcmd:1} completes the proof.

\subsection{Proof of Lemma~\ref{lem:ocmd}}
Firstly, by using the convexity of $r(\cdot)$, for all $\z\in\mathcal{U}$ we have
\begin{align}
    {}&\left\langle \w - \z, \gamma\eta \right\rangle + r(\w)-r(\z) \nonumber\\
    ={}& \left\langle \w - \z, \gamma\eta \right\rangle + r(\w)-r(\z_+) + r(\z_+) - r(\z)\nonumber\\
    \leq{}& \left\langle \w - \z, \gamma\eta \right\rangle + \left\langle \w - \z_+, \nabla r(\w) \right\rangle + \left\langle \z_+ - \z, \nabla r(\z_+) \right\rangle\nonumber\\
    ={}&\left\langle \w-\z_+,\gamma\eta -\gamma\xi  \right\rangle + \underbrace{\left\langle \w-\z_+,\gamma\xi+\nabla r(\w) \right\rangle}_{\ta} + \underbrace{\left\langle  \z_+ - \z,\gamma\eta+\nabla r(\z_+)\right\rangle}_{\tb}.\label{linearloss}
\end{align}
Next we introduce the following lemma   \citep[Lemma 3.1]{TAC:2023:Scroccaro} to bound $\ta$ and $\tb$.
\begin{lemma}\label{lemma3.1ofScroccaro}
    Suppose that $\X$ is a closed convex set. Let $\varphi:\X \rightarrow \R$ be a convex function and $\eta > 0$. Define
    \begin{align*}
        u = \argmin_{x\in\X}\{\eta \varphi(x)+\B_\omega(x,v)\}.
    \end{align*}
    It follows that, for all $z \in \X$ and $g(u)\in \partial\varphi(u)$,
    \begin{align*}
        \eta\langle g(u),u-z \rangle \leq \B_\omega(z,v) - \B_\omega(z,u) - \B_\omega(u,v).
    \end{align*}
\end{lemma}
Applying Lemma~\ref{lemma3.1ofScroccaro} to the update rules, we have
\begin{align*}
    &\ta = \left\langle \gamma\xi + \nabla r(\w),\w - \z_+ \right\rangle \leq \B_{\omega}(\z_+,\z_{-}) - \B_{\omega}(\z_+,\w)-\B_{\omega}(\w,\z_{-}),\\
    &\tb = \left\langle \gamma\eta + \nabla r(\z_+),\z_+ - \z \right\rangle \leq \B_{\omega}(\z,\z_{-}) - \B_{\omega}(\z,\z_+)-\B_{\omega}(\z_+,\z_{-}).
\end{align*}
As a result, we can bound~\eqref{linearloss} as
\begin{align*}
    {}&\left\langle \w - \z, \gamma\eta \right\rangle + r(\w)-r(\z) \\
    \leq{}& \left\langle \w-\z_+,\gamma\eta -\gamma\xi  \right\rangle + \B_{\omega}(\z,\z_{-}) - \B_{\omega}(\z,\z_+) - \B_{\omega}(\z_+,\w)-\B_{\omega}(\w,\z_{-})\\
    \leq{}&\gamma\left\|\eta -\xi\right\|_*\left\|\w-\z_+\right\|+ \B_{\omega}(\z,\z_{-}) - \B_{\omega}(\z,\z_+) - \B_{\omega}(\z_+,\w)-\B_{\omega}(\w,\z_{-})\\
    \leq{}&\gamma\left\|\eta -\xi\right\|_*\left\|\w-\z_+\right\|+ \B_{\omega}(\z,\z_{-}) - \B_{\omega}(\z,\z_+)- \frac{\alpha}{2} \big[ \|\w - \z_+ \|_2^2 + \| \z_- - \w \|_2^2 \big],
\end{align*}
where the last inequality is due to that $\omega(\cdot)$ is $\alpha$-strongly convex. Then, we exploit Lemma 3.2 of~\citet{TAC:2023:Scroccaro} to get
\begin{align*}
    \left\|\w-\z_+\right\| \leq \alpha^{-1}\gamma\|\xi -\eta\|_*.
\end{align*}
Hence, we finally arrive at
\begin{align*}
    {}&\left\langle \w - \z, \gamma\eta \right\rangle + r(\w)-r(\z) \\
    \leq{}& \frac{\gamma^2}{\alpha}\|\xi -\eta\|_* + \B_{\omega}(\z,\z_{-}) - \B_{\omega}(\z,\z_+)- \frac{\alpha}{2} \big[ \|\w - \z_+ \|_2^2 + \| \z_- - \w \|_2^2 \big].
\end{align*}

\vskip 0.2in
%\bibliography{E:/MyPaper/ref}
\bibliography{ref}

\end{document}